\documentclass{article}

\usepackage[final]{neurips_2025}

\usepackage[utf8]{inputenc} %
\usepackage[T1]{fontenc}    %
\usepackage{hyperref}       %
\usepackage{url}            %
\usepackage{booktabs}       %
\usepackage{amsfonts}       %
\usepackage{nicefrac}       %
\usepackage{xcolor}         %

\usepackage{multirow}
\usepackage{mathtools}
\usepackage{natbib}
\usepackage{booktabs}
\usepackage{array}
\usepackage{makecell}
\usepackage{caption}
\usepackage{amsmath}
\usepackage{amsthm}
\usepackage{comment}
\usepackage{graphicx}
\usepackage[capitalise]{cleveref}

\usepackage[ruled,linesnumbered,noend]{algorithm2e}
\usepackage{setspace}
\usepackage[noend]{algpseudocode}
\SetKwInput{KwInput}{Input}
\SetKwInput{KwOutput}{Output}
\SetKwInput{KwHyperparameters}{Hyperparameters}

\ifpdf
\crefname{algocf}{algorithm}{algorithms}

\fi

\definecolor{provercolor}{RGB}{192, 35, 55}
\definecolor{verifiercolor}{RGB}{78, 167, 46}

\algdef{SE}[SUBALG]{Indent}{EndIndent}{}{\algorithmicend\ }%
\algtext*{Indent}
\algtext*{EndIndent}

\newtheorem{theorem}{Theorem}[section]       
\newtheorem{definition}[theorem]{Definition}
\newtheorem{remark}[theorem]{Remark}
\newtheorem{lemma}[theorem]{Lemma}

\newtheorem{fact}[theorem]{Fact}

\algrenewcommand{\algorithmiccomment}[1]{\hfill \textcolor{orange}{\texttt{\# #1}}}

\newcommand{\N}{\mathbb{N}}
\newcommand{\Z}{\mathbb{Z}}
\newcommand{\R}{\mathbb{R}}

\newcommand{\defeq}{\coloneqq}
\renewcommand{\gets}{\sim}
\DeclareMathOperator*{\argmax}{\mathrm{argmax}}
\DeclareMathOperator*{\E}{\mathbb{E}}

\renewcommand{\epsilon}{\varepsilon}

\newcommand{\Tok}{\Sigma}
\newcommand{\Seq}{\Tok^\ast}
\newcommand{\accepts}{\mathrm{ accepts}}

\newcommand{\seq}[1]{\mathbf{#1}}
\newcommand{\eps}{\varepsilon}

\newcommand{\ver}{\mathrm{ver}}
\newcommand{\agree}{A}
\newcommand{\Transcript}{\mathcal{T}}

\newcommand{\Acc}{\mathrm{Acc}}
\newcommand{\vp}[2]{\left<#1,#2\right>}
\newcommand{\Bnorm}{B_{\mathrm{Norm}}}
\newcommand{\Blip}{B_{\mathrm{Lip}}}
\newcommand{\Cprov}{C}

\def\draft{1}   %

\newcommand{\authnote}[3]{\textcolor{#3}{[{\footnotesize {\bf #1:} { {#2}}}]}}

\ifnum\draft=0
\renewcommand{\authnote}[3]{}  %
\fi

\newcommand{\remove}[1]{}

\newcommand{\captitle}[1]{\textbf{{#1}}}

\newcommand{\codeurl}{https://github.com/orrp/self-proving-models}

\title{Models That Prove Their Own Correctness}

\author{Noga Amit\thanks{Authors listed alphabetically. This paper appeared under the title ``A Theory for Worst-Case vs. Average-Case Guarantees for LLMs'' \citep{neurips}.} \\
    UC Berkeley \\
    \And
    Shafi Goldwasser\footnotemark[1]{} \\
  UC Berkeley \\
  \And
  Orr Paradise\footnotemark[1]{} \\
  UC Berkeley  \\
   \And
   Guy N. Rothblum\footnotemark[1]{} \\
   Apple \\
}

\begin{document}

\maketitle

\begin{abstract}
How can we trust the correctness of a learned model on a particular input of interest? Model accuracy is typically measured \emph{on average} over a distribution of inputs, giving no guarantee for any fixed input.
This paper proposes a theoretically-founded solution to this problem: to train \emph{Self-Proving models} that prove the correctness of their output to 
a verification algorithm $V$ via an Interactive Proof.
Self-Proving models satisfy that, with high probability over an input sampled from a given distribution, the model generates a correct output \emph{and} successfully proves its correctness to $V\!$. The \emph{soundness} property of $V$ guarantees that, for \emph{every} input, no model can convince $V$ of the correctness of an incorrect output. Thus, a 
Self-Proving model proves correctness of most of its outputs, while \emph{all} incorrect outputs (of any model) are detected by $V$. We devise and analyze two generic methods for learning 
Self-Proving models: \emph{Transcript
Learning (TL)} which relies on access to transcripts of accepting interactions, and \emph{Reinforcement Learning from Verifier Feedback (RLVF)} which trains a model by emulating interactions with the verifier.
\end{abstract}  

\tableofcontents

\section{Introduction}
\begin{figure}[t]
   \centering
   \includegraphics[width=0.7\columnwidth]{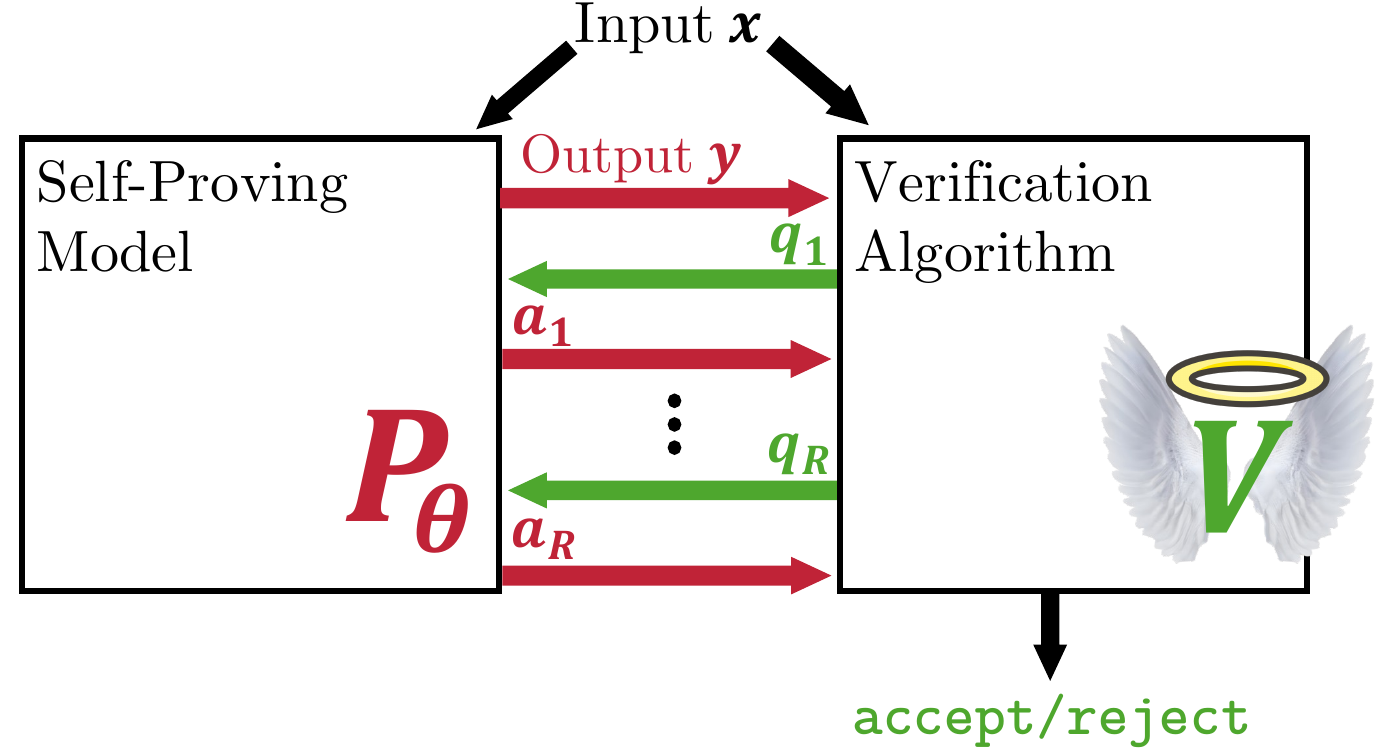}
   \caption{\textbf{Self-Proving models.} For input $x$, Self-Proving model $P_\theta$ generates an output $y$ and sends it to a Verification Algorithm $V$. Then, over $i \in [R]$ rounds, $V$ sends query $q_i$, and receives an answer $a_i$ from $P_\theta$. Finally, $V$ decides (``accept/reject'') whether it is convinced that $y$ is a correct output for $x$.}
   \label{fig:main}
\end{figure}
Bob is studying for his algebra exam and stumbles upon a question $Q$ that he cannot solve. He queries a Large Language Model (LLM) for the answer, and it responds with a number: 42. Bob is aware of recent research showing that the LLM attains a 90\% score on algebra benchmarks (cf. \citealt{FriederPCGSLPB23}), but should he trust that the answer to his particular question $Q$ is indeed 42? 

Bob could ask the LLM to explain its answer in natural language. Though he must proceed with caution, as the LLM might try to convince him of an incorrect answer \citep{TurpinMPB23}. Moreover, even if 42 is the correct answer, the LLM may fail to produce a convincing proof \citep{WangY023}. If only the LLM could formally prove its answer, Bob would verify the proof and be convinced.

This paper initiates the study of \emph{Self-Proving models} (\cref{fig:main}) that prove the correctness of their answers via an Interactive Proof system \citep{GoldwasserMR85}. Self-Proving models successfully convince a verification algorithm $V$ with \emph{worst-case soundness guarantees}: for any question, $V$ rejects all incorrect answers with high probability over the interaction. This guarantee holds even against provers that have access to $V$'s specification, and unbounded computational power.

Our contributions are as follows.
\begin{itemize}
	\item We define Self-Proving models (\Cref{sec:defs}).
	\item We propose two methods for learning Self-Proving models (\Cref{sec:learning_ver_mod}). The first, \emph{Transcript Learning (TL)}, relies on access to transcripts of accepting interactions. 
    The second method, \emph{Reinforcement Learning from Verifier Feedback (RLVF)}, trains a model by emulating interactions with the verifier. 
    \item We prove gradient approximation lemmas for both methods (\Cref{lem:grad_restated,lem:grad_rlvf}), and a convergence bounds for TL under convexity and Lipschitzness assumptions (\Cref{sec:converge_TL}). These are supplemented by empirical validation on a simple arithmetic capability (\Cref{apx:experiments}). Code and data are at \url{\codeurl}
\end{itemize}

This paper develops a theory of learned models that prove their own correctness via an interactive proof system, and thus lies at the intersection of machine learning and Interactive Proof systems. We defer the discussion of relevant prior work from these areas to the related work section in \Cref{sec:related_work}.
The rich and well-studied question of \emph{which} settings are verifiable within an interactive proof system is beyond our scope. Our theory is general in that it applies to \emph{any} setting which is verifiable within an interactive proof system, e.g., any decision problem solvable in polynomial space \citep{Shamir92}. For a broader introduction to proof systems, see \citet{Goldreich08primer}.

\section{Related Work}\label{sec:related_work}
We overview related work from machine learning (ML) and interactive proof systems (IPs) literature.

\paragraph{ML and IPs.}
IPs have found numerous applications in ML towards a diverse set of goals.
\citet{AnilZWG21} introduce Prover--Verifier Games (PVGs), a game-theoretic framework for learned provers and learned verifiers. Since our paper initially appeared, PVGs were further investigated in at least two subsequent works: \citet{HammondD24} study multi-prover and Zero Knowledge variants of PVGs. Additionally, \citet{KirchnerCELMB24} successfully utilize PVGs towards obtaining human-legible outputs from LLMs. Notably, they require a relaxed completeness guarantee of their learned proof system---this requirement is the same as our \Cref{def:ver_mod} of Self-Proving models.

Beyond PVGs, \citet{Waldchen24a} cast the problem of model interpretability as a Prover--Verifier interaction between a learned feature selector and a learned feature classifier. Debate systems \citep{CondonFLS95}, a multiprover variant of IPs, were considered for aligning models with human values \citep{IrvingCA18,BrownIP23}. In such Debate systems, two competing models are each given an alleged answer $y \neq y'$, and attempt to prove the correctness of their answer to a (human or learned) judge.
Lastly, \citet{MurtyPS23} define Pseudointelligence: a model learner $L_M$ and an evaluator learner $L_E$ are each given samples from a ground-truth; $L_M$ learns a model of the ground-truth, while $L_E$ learns an evaluator of such models; the learned evaluator then attempts to distinguish between the learned model and the ground-truth in a Turing Test-like interaction.

All of these works consider \emph{learned verifiers}, whereas our work focuses on training models that interact with a manually-defined verifier. More related in this regard is IP-PAC \citep{GoldwasserRSY21}, in which a learner proves that she learned a model that is Probably Approximately Correct \citep{Valiant84}. We, however, consider \emph{models} that prove their own correctness on a \emph{per-input basis}, rather than \emph{learners} that prove \emph{average-case correctness} of a model.

\paragraph{Models that generate formal proofs.}
Self-Proving models are verified by an algorithm with formal completeness and soundness guarantees (see \Cref{def:sys-ver}). In this sense, Self-Proving models generate a formal proof of the correctness of their output. Several works propose specialized models that generate formal proofs.

AlphaGeometry \citep{TrinhWLHL24} is capable of formally proving olympiad-level geometry problems; Others have trained models to produce proofs in \citet{GransdenWR15,PoluS20} and others train models to produce proofs in Coq \citep{GransdenWR15}, Metamath \citep{PoluS20}, Lean \citep{YangSGCSYGPA23}, or manually-defined deduction rules \citep{TafjordDC20}; FunSearch \citep{Romera24} evolves LLM-generated programs by systematically evaluating their correctness. Indeed, all of these can be cast as Self-Proving models developed for \emph{specific proof systems}. Meanwhile, this work defines and studies the class of such models \emph{in general}. Several works (e.g. \citealt{WelleckLLHC22}) consider models that generate natural language proofs or explanations, which are fundamentally different from formal proofs (or provers) verified by an algorithm.

\paragraph{Training on intermediate steps.} Chain-of-Though (CoT, \citealt{Wei0SBIXCLZ22}) refers to additional supervision on a model in the form of intermediate reasoning steps. CoT is known to improve model performance whether included in-context \citep{Wei0SBIXCLZ22} or in the training phase itself \citep{YangSAN22}. Transcript Learning (TL, \Cref{sec:ver_base}) can be viewed as training the model on a Chain-of-Thought induced by the interaction of a verifier and an honest prover (\Cref{def:sys-ver}).

To complete the analogy, let us adopt the terminology of \citet{Uesato22}, who consider \emph{outcome supervision} and \emph{process supervision}. In our case, the \emph{outcome} is the decision of the verifier, and the \emph{process} is the interaction between the verifier and the model. Thus, Reinforcement Learning from Verifier Feedback (RLVF, \Cref{sec:ver_amp}) is outcome-supervised while TL is process-supervised. In a recent work, \citet{LightmanKBEBLLSSC24} find that process-supervised transformers outperform outcome-supervised ones on the MATH dataset \citep{HendrycksBKABTS21}.

\paragraph{Subsequent work on RLVF.}
Following the preprint version of this paper on May 2024, Reinforcement Learning from Verifier Feedback (RLVF, \Cref{sec:ver_amp}) has been implemented and widely adopted. Of particular note is RLVR (Reinforcement Learning from Verifiable Reward, \citealt{LambertEtAl24}), which implements RLVF by adding KL regularization to the RL objective of \Cref{alg:RLVF}. This implementation provided valuable empirical validation and state-of-the-art performance on full-scale LLMs, sparked additional empirical analysis \citep{WuXLHC25,zhou2025rl}, and led to widespread adoption across numerous domains including medical reasoning, computer vision, and robotics \citep{med-rlvf,vision-rlvf,robotics-rlvf}. Most recently, \cite{PyatkinEtAl25} presented IFBench which measures performance in interactive (3-round) proof systems. These empirical contributions complement our theoretical foundations, demonstrating the surprising power and applicability of RLVF as a practical post-training method.

\section{Defining Self-Proving Models}\label{sec:defs}
We introduce and formally define our learning framework in which models prove the correctness of their output. We start with preliminaries from the learning theory and proof systems literatures in \Cref{sec:prelim}. We then introduce our main definition in \cref{sec:ver_mod}.

\subsection{Preliminaries}\label{sec:prelim}

Let $\Sigma$ be a finite set of tokens and $\Sigma^\ast$ denote the set of finite sequences of such tokens. We consider sequence-to-sequence models $F_\theta \colon \Seq \to \Seq$, which are total functions that produce an output for each possible input sequence. A model is parameterized by a real-valued, finite dimensional vector $\theta$. We consider models as \emph{randomized} functions, meaning that $F_\theta(x)$ is a random variable over $\Seq$, of which samples are denoted by $y \gets F_\theta(x)$.

Before we can define models that prove their own correctness, we must first define correctness. Correctness is defined with respect to an input distribution $\mu$ over $\Seq$, and a ground-truth $F^\ast$ that defines correct answers.
For simplicity of presentation, we focus on the case that each input $x \in \Seq$ has exactly one correct output $F^\ast(x) \in \Seq$, and a zero-one loss function on outputs (the general case is deferred to \Cref{apx:defs-gen}). The fundamental goal of machine learning can be thought of as learning a model of the ground-truth $F^\ast$. Formally,
\begin{definition}[Correctness]\label{def:correctness}
	Let $\mu$ be a distribution of input sequences in $\Seq$ and let $F^\ast \colon \Seq \to \Seq$ be a fixed (deterministic) ground-truth function. For any $\alpha \in [0,1]$, we say that model $F_\theta$ is $\alpha$-correct (with respect to $\mu$) if
	\begin{equation*}
		\Pr_{\substack{x \gets \mu \\ y \gets F_\theta(x)}}[y = F^\ast(x)] \geq \alpha.
	\end{equation*}
\end{definition}

An \emph{interactive proof system} \citep{GoldwasserMR85} is a protocol carried out between an efficient \emph{verifier} and a computationally unbounded \emph{prover}. The prover attempts to convince the verifier of the correctness of some assertion, while the verifier accepts only correct claims. The prover is powerful yet untrusted; in spite of this, the verifier must reject false claims with high probability.

In the context of this work, it is important to note that the verifier is \emph{manually-defined} (as opposed to learned). Formally, the verifier is a probabilistic polynomial-time algorithm tailored to a particular ground-truth capability $F^\ast$. Informally, the verifier is the anchor of trust: think of the verifier as an efficient and simple algorithm, hosted in a trustworthy environment.

Given an input $x \in \Seq$, the model $F_\theta$ ``claims'' that $y \gets F_\theta(x)$ is correct. We now define what it means to \emph{prove} this claim. We will use $P_\theta$ to denote Self-Proving models, noting that they are formally the same object\footnote{Both are randomized mappings from $\Seq$ to $\Seq$.} as non-Self-Proving (``vanilla'') models $F_\theta$. This notational change is to emphasize that $P_\theta$ first outputs $y \gets P_\theta(x)$ \emph{and is then prompted by the verifier}, unlike $F_\theta$ who only generates an output $y \gets F_\theta(x)$.

A Self-Proving model proves that $y \gets P_\theta(x)$ is correct to a verifier $V$ over the course of $R$ rounds of interaction (\Cref{fig:main}). In each round $i \in [R]$, verifier $V$ queries $P_\theta$ on a sequence $q_i \in \Seq$ to obtain an answer $a_i \in \Seq$; once the interaction is over, $V$ accepts or rejects. For fixed $x, y \in \Seq$, the decision of $V$ after interacting with $P_\theta$ is a random variable over $V$'s decision (accept/reject), determined by the randomness of $V$ and $P_\theta$. The decision random variable is denoted by $\vp{V}{P_\theta}(x,y)$.

Next, we present a definition of Interactive Proofs restricted to our setting.

\begin{definition}\label{def:sys-ver}
    Fix a soundness error $s \in (0,1)$, a finite set of tokens $\Sigma$ and a ground-truth $F^\ast \colon \Seq \to \Seq$. A \emph{verifier $V$ (in an Interactive Proof) for $F^\ast$} is a probabilistic polynomial-time algorithm that is given explicit inputs $x,y \in \Seq$ and black-box (oracle) query access to a prover $P$.\footnote{We intentionally write $P$ rather than $P_\theta$: Interactive Proofs are defined with respect to all possible provers, not just parameterized ones.} It \emph{interacts} with $P$ over $R$ rounds (see \Cref{fig:main}) and outputs a decision $\vp{V}{P}(x,y)\in \{\mathrm{reject}, \mathrm{accept} \}$. Verifier $V$ satisfies the following two guarantees:
    \begin{itemize}
        \item \emph{Completeness:} There exists an \emph{honest prover} $P^\ast$ such that, for all $x \in \Seq$,
        \begin{equation*}
        	\Pr[\vp{V}{P^\ast}\!(x,F^\ast(x))\ \accepts] = 1,
        \end{equation*}
        where the probability is over the randomness of $V$.\footnote{WLOG, the honest prover is deterministic by fixing the optimal randomness of a randomized prover.}
        \item \emph{Soundness:} For all $P$ and for all $x, y \in \Seq$, if $y \neq F^\ast(x)$ then
        \begin{equation*}
        	\Pr[\vp{V}{P}(x,y)\ \accepts] \leq s,
        \end{equation*}
        where the probability is over the randomness of $V$ and $P$, and $s$ is the soundness error.
    \end{itemize}
\end{definition}
The efficiency of an interactive proof is usually measured with respect to four parameters: the round complexity $R$, the communication complexity (the overall number of bits transferred during the interaction), $P^\ast$'s efficiency and $V$'s efficiency. 
These complexity measures scale with the computational complexity of computing the ground-truth $F^\ast$. For example, an interactive proof for a complex $F^\ast$ may require multiple rounds of interaction.

\begin{remark}[Verifier efficiency]\label{rem:verifier_efficiency}
\cref{def:sys-ver} requires that $V$ is a polynomial-time algorithm whereas provers are unbounded. This captures a requirement for \emph{efficient verification}. We chose polynomial time as a measure of efficiency because it is common in the Proof systems literature. That said, one could adapt \Cref{def:sys-ver} to fit alternative efficiency measures, such as space complexity \citep{CondonL89} or circuit depth \citep{GoldwasserGHKR07}. Regardless of which measure is taken, to avoid a trivial definition it is crucial that $V$ should be more efficient than the honest prover $P^\ast$; else, $V$ can simply execute $P^\ast$ to perform the computation itself.
\end{remark}

By definition, the soundness error $s$ of a verifier $V$ bounds the probability that it is mistakenly convinced of an incorrect output; in that sense, the smaller $s$, the ``better'' the verifier $V$. In our setting, we think of a manually-defined verifier $V$ who is formally proven (by a human) to have a small soundness error by analysis of $V$'s specification.

As depicted in \Cref{fig:main}, each of the model's answers depends on all previous queries and answers in the interaction. This captures the setting of \emph{stateful models}, e.g. a session with a chatbot.

Towards defining Self-Proving models (\Cref{sec:ver_mod}), let us observe the following. Completeness and soundness are \emph{worst-case guarantees}, meaning that they hold for all possible inputs $x\in \Seq$. In particular, completeness implies that for all $x \in \Seq$, the honest prover $P^\ast$ convinces $V$ of the correctness of $F^\ast(x)$; in classical proof systems there is no guarantee that an ``almost honest'' prover can convince the verifier (cf. \citealt{Paradise21}). Yet, if we are to \emph{learn} a prover $P_\theta$, we cannot expect it to agree with $P^\ast$ perfectly, nor can we expect it to always output $F^\ast(x)$. Indeed, Self-Proving models will have a \emph{distributional guarantee} with respect to inputs $x \gets \mu$. This distinction is summarized in \Cref{tab:guarantees}.

\begin{table}
  \caption{\textbf{Formal guarantees.} Completeness and soundness are fundamental guarantees of a verification algorithm $V$. Verifiability (novel in this work) is a feature of a model $P_\theta$ with respect to a verifier $V$ and input distribution $\mu$. Importantly, $V$'s soundness holds for any input $x$ and output $y$.}
  \label{tab:guarantees}
  \centering
  \begin{tabular}{llll}
    \toprule
     & Guarantee & Type & Def. \\
    \midrule
    {\color{verifiercolor}$V$} 
      & Completeness \& Soundness 
      & Worst-case: $\forall x,y$ 
      & \ref{def:sys-ver} \\
      
    {\color{provercolor}$P_\theta$} 
      & Verifiability 
      & Average-case: $x \sim \mu,\ y \sim P_\theta(x)$ 
      & \ref{def:ver_mod} \\
    \bottomrule
  \end{tabular}
\end{table}

\subsection{Self-Proving Models}\label{sec:ver_mod}
We define the \emph{Verifiability} of a model $P_\theta$ with respect to an input distribution $\mu$ and a verifier $V$. Intuitively, Verifiability captures the ability of the model to prove the correctness of its answer $y \gets P_\theta(x)$, when the input $x$ is sampled from $\mu$. We refer to models capable of proving their own correctness as \emph{Self-Proving models}. Notice that, as in \cref{def:sys-ver}, the verifier is fixed and agnostic to the choice of the Self-Proving model.

\begin{definition}[Self-Proving model]\label{def:ver_mod}
	Fix a verifier $V$ for a ground-truth $F^\ast \colon \Seq \to \Seq$ as in \cref{def:sys-ver}, and a distribution $\mu$ over inputs $\Seq$. The \emph{Verifiability} of a model $P_\theta\colon \Seq \to \Seq$ is defined as
	\begin{equation}\label{eq:ver_mod}
		\ver_{V,\mu}(\theta) \defeq \Pr_{\substack{x \gets \mu \\ y \gets P_\theta(x)}}\left[
		\vp{V}{P_\theta}(x,y)\  \accepts
		\right].
	\end{equation}
	We say that model $P_\theta$ is \emph{$\beta$-Self-Proving} with respect to $V$ and $\mu$ if $\ver_{V,\mu}(\theta) \geq \beta$.
\end{definition}
\begin{remark}[Verifiability $\implies$ correctness]\label{rem:spm-implies-correctness}
	Notice that the ground-truth $F^\ast$ does not appear in \Cref{def:ver_mod} except for the first sentence. Indeed, once it is established that $V$ is a verifier for $F^\ast$ (as per \Cref{def:sys-ver}), then Verifiability w.r.t $V$ implies correctness w.r.t $F^\ast$: Consider any input distribution $\mu$, ground-truth $F^\ast$, and a verifier $V$ for $F^\ast$ with soundness error $s$. By a union bound, if a model $P_\theta$ is $\beta$-Verifiable, then it is $(\beta-s)$-correct. That is to say, Verifiability is formally a stronger guarantee than correctness when $V$ has small soundness error $s$.
\end{remark}
As depicted in \Cref{fig:main}, a Self-Proving model $P_\theta$ plays a dual role: first, it generates an output $y \gets P_\theta(x)$, and then it proves the correctness of this output to $V$. Note also that Verifiability is a feature of a \emph{model}, unlike completeness and soundness which are features of a \emph{verifier} (see \Cref{tab:guarantees}).

The benefit of Verifiability over correctness is captured by the following scenario. Alice wishes to use a model $P_\theta$ to compute some functionality $F^\ast$ on an input $x_0$ in a high risk setting. Alice generates $y_0 \gets P_\theta(x_0)$. Should Alice trust that $y_0$ is correct? If Alice has a held-out set of labeled samples, she can estimate $P_\theta$'s average correctness on $\mu$. Unfortunately, (average) correctness provides no guarantee regarding the correctness of the particular $(x_0,y_0)$ that Alice has in hand. If, however, Alice has access to a verifier $V$ for which $P_\theta$ is Self-Proving, then she can trust the model on an input-by-input (rather than average-case) basis: Alice can execute $V$ on $(x_0,y_0)$ and black-box access to $P_\theta$. Soundness of $V$ guarantees that if $y_0$ is incorrect, then $V$ rejects with high probability, in which case Alice should either generate $P_\theta(x_0)$ again---or find a better model.

\subsection{A Definition for General Loss Functions and One-To-Many Relations}\label{apx:defs-gen}

We present variants of Self-Proving models (\Cref{def:ver_mod}) generalized to one-to-many relations, and general bounded loss functions. While these generalizations provide a richer framework that may accommodate a wider range of applications, the theorems in this paper are based on the forgoing \Cref{def:ver_mod}, which captures the essential properties while remaining mathematically manageable.

\paragraph{General (bounded) loss functions.}
In \Cref{def:correctness} we implicitly use the 0-1 loss when measuring the correctness of a model: For any $x \in X$, we measure only whether the model generated the correct output $y = F^\ast(x)$, but not how ``far'' the generated $y$ was from $F^\ast(x)$. It is often the case in machine learning that we would be satisfied with models that generate a ``nearly-correct'' output. This is formalized by specifying a loss function $\ell \colon \Seq \times \Seq \to [0, 1]$ and measuring the probability that $\ell(x, y)$ is smaller than some threshold $\lambda \in [0, 1)$, where $x$ is drawn from the input distribution $\mu$, and $y$ is generated by the model when given input $x$.

In the context of language modeling, different loss function allow for a more fine-grained treatment of the \emph{semantics} of a given task. As an example, consider the \emph{prime-counting task}:
\begin{itemize}
	\item Given an integer $x < 10^9$, output the number of primes less than or equal to $x$.
\end{itemize}
In the notation of \Cref{sec:defs}, the prime-counting task would be captured by the ground-truth function
\begin{equation*}
	F^\ast(x) \defeq \left| \left\{ p \in \N\ \middle|\ p \leq x,\ p\textrm{ is prime} \right\} \right|.\footnote{Formally, the input and output are strings in $\Seq$ representing integers (e.g. in decimal representation).}
\end{equation*}
Per \Cref{def:correctness}, any output other than $F^\ast(x)$ is ``just as incorrect'' as any other. Yet, we might prefer outputs that are closer to the correct answer, say, in $L_1$ norm. This preference can be captured by the following bounded loss function
\begin{equation*}
	\ell_1(x,y) \defeq \begin{cases}
		|y - F^\ast(x)| \cdot 10^{-9} \quad &\textrm{if }y \leq 10^9\\
		1 &\textrm{else.}
	\end{cases}
\end{equation*}
In particular, if we are interested in knowing the answer only up to some additive constant $C$, we could say that an output $y$ is ``correct-enough'' if $\ell_1(x,y) \leq C \cdot 10^{-9}$.

More generally, we relax \Cref{def:correctness} to capture approximate correctness as follows.
\begin{definition}[Approximate correctness]\label{def:apx-correctness}
		Let $\mu$ be a distribution over input sequences in $\Seq$ and let $\ell\colon \Seq \times \Seq \to [0,1]$ be a loss function. For any $\alpha, \lambda \in [0,1]$, we say that model $F_\theta$ is $(\alpha,\lambda)$-correct with respect to $\mu$ if
	\begin{equation*}
		\Pr_{\substack{x \gets \mu \\ y \gets F_\theta(x)}}[\ell(x, y) \leq \lambda] \geq \alpha.
	\end{equation*}
\end{definition}

\paragraph{One-to-many-relations.}
In \Cref{sec:defs}, we focused on the setting of models of a ground-truth function $F^\ast \colon \Seq \to \Seq$. That is, when each input $x$ has exactly one correct output, namely $F^\ast(x)$. A more general setting would be to consider a ground-truth \emph{relation} $L \subseteq \Seq \times \Seq$. Then, we say that $y$ is a correct output for $x$ if $(x, y) \in L$. Importantly, this allows a single $x$ to have many possible correct outputs, or none at all.

Note that we must take care to choose a loss function $\ell$ that captures correctness with respect to the relation $L$, i.e., $\ell(x,y)=0$ if and only if $(x, y) \in L$.
Equivalently, any loss function $\ell$ induces a relation $L \defeq \{ (x,y) \ |\  \ell(x,y) = 0 \}$. Therefore, our relaxation to approximate-correctness \Cref{def:apx-correctness} already captures the setting of one-to-many relations, since an input $x$ may have multiple $y^\ast$ such that $\ell(x,y^\ast)=0$.

\subsubsection{The General Definition}
We first present a relaxed definition of Interactive Proof systems for verifying approximate-correctness.
\begin{definition}[\Cref{def:sys-ver}, generalized]\label{def:sys-ver-gen}
    Fix a soundness error $s \in (0,1)$, a threshold $\lambda \in [0,1)$, a finite set of tokens $\Sigma$, and a loss function $\ell \colon \Seq \times \Seq \to [0,1]$. A \emph{verifier $V$ for $\ell$ with threshold $\lambda$} is a probabilistic polynomial-time algorithm that is given explicit inputs $x,y \in \Seq$ and black-box (oracle) query access to a prover $P$. It \emph{interacts} with $P$ over $R$ rounds (see \Cref{fig:main}) and outputs a decision $\vp{V}{P}(x,y)\in \{\mathrm{reject}, \mathrm{accept} \}$. Verifier $V$ satisfies the following two guarantees:
    \begin{itemize}
        \item \emph{Completeness:} There exists an \emph{honest prover} $P^\ast$ such that, for all $x, y \in \Seq$, if $\ell(x,y) = 0$ then
        \begin{equation*}
        	\Pr[\vp{V}{P^\ast}\!(x,y)\ \accepts] = 1,
        \end{equation*}
        where the probability is over the randomness of $V$.
        \item \emph{Soundness:} For all $P$ and for all $x, y \in \Seq$, if $\ell(x, y) > \lambda$ then
        \begin{equation*}
        	\Pr[\vp{V}{P}(x,y)\ \accepts] \leq s,
        \end{equation*}
        where the probability is over the randomness of $V$ and $P$, and $s$ is the soundness error.
    \end{itemize}
\end{definition}
Indeed, for a given ground-truth function $F^\ast \colon \Seq \to \Seq$, \Cref{def:sys-ver} can be recovered by choosing the 0-1 loss
\begin{equation*}
	\ell_{F^\ast}(x,y) \defeq \begin{cases}
		1\quad \mathrm{if }\  x \neq F^\ast(y) \\
		0\quad \mathrm{else}
	\end{cases}
\end{equation*}
and any threshold $\lambda \in [0,1)$.

\begin{remark}[Connection to Interactive Proofs of Proximity]
\Cref{def:sys-ver-gen} can be seen as a slight generalization of (perfect completeness) Interactive Proofs of Proximity (IPPs, \citealt{RothblumVW13}). An IPP for a relation $L \subseteq \Seq \times \Seq$ with proximity parameter $\lambda$ is obtained by instantiating \Cref{def:sys-ver-gen} with the loss function $\ell_{\mathrm{Hamming}}$ defined by
\begin{equation*}
	\ell_{\mathrm{Hamming}}(x,y) \defeq	\min \left\{ \frac{\#\{i\ |\ y_i \neq y^\ast_i \}}{|y|} \ \middle|\ (x,y^\ast)\in L,\ |y^\ast| = |y| \right\},
\end{equation*}
that is, $\ell_{\mathrm{Hamming}}(x,y)$ is the fraction of tokens in $y$ that must be changed to obtain an output $y^\ast$ with $(x,y^\ast) \in L$.
However, the motivation of \cite{RothblumVW13} was studying sublinear time verification, whereas ours is to relax the requirements of traditional Interactive Proofs towards meeting common desiderata in machine learning.
\end{remark}

With this relaxed notion of Interactive Proofs in hand, we are now ready to define Self-Proving models for general (bounded) loss functions.
\begin{definition}[\Cref{def:ver_mod}, generalized]\label{def:ver_mod_gen}
	Fix a loss function $\ell \colon \Seq \times \Seq \to [0,1]$, a verifier $V$ for $\ell$ with threshold $\lambda \in [0,1)$ as in \Cref{def:sys-ver-gen}, and a distribution $\mu$ over inputs $\Seq$.
	The \emph{Verifiability} of a model $P_\theta \coloneq \Seq \to \Seq$ is defined as 
	\begin{equation*}
		\ver_{V,\mu}(\theta) \defeq \Pr_{\substack{x \gets \mu \\ y \gets P_\theta(x)}}\left[
		\vp{V}{P_\theta}(x,y)\  \accepts
		\right].
	\end{equation*}
	We say that model $P_\theta$ is \emph{$\beta$-Self-Proving} with respect to $V$ and $\mu$ if $\ver_{V,\mu}(\theta) \geq \beta$.
\end{definition}
Analogously to \Cref{rem:spm-implies-correctness}, we observe that Verifiability as per \Cref{def:ver_mod_gen} implies approximate-correctness: Suppose $P_\theta$ is $\beta$-Self-Proving model with respect to a verifier $V$ that has soundness error $s$ and threshold parameter $\lambda$ for loss function $\ell$. Then, by a union bound,
\begin{equation*}
	\Pr_{\substack{x \gets \mu \\ y \gets P_\theta(x)}}\left[
		\ell(x,y) \leq \lambda
		\right] \geq \beta - s.
\end{equation*}
Importantly, as emphasized throughout this paper, soundness of $V$ implies that for \emph{all} inputs $x$, any output $y$ such that $\ell(x, y) > \lambda$ is rejected with high probability ($1-s$).

\section{Algorithms for Learning Self-Proving Models}\label{sec:learning_ver_mod}
With a sound verifier $V$ at hand, obtaining Self-Proving models with respect to $V$ holds great promise: a user that prompts the model with input $x$ does not need to take it on good faith that $P_\theta(x)$ is correct; she may simply verify this herself by executing the verification protocol.
How, then, can we learn models that are not just approximately-correct, but Self-Proving as well?

We focus on differentiable autoregressive models, and assume that the learner has access to input samples $x \gets \mu$ and correct outputs $F^\ast(x)$, as well as the verifier's specification (code). Additionally, the learner can emulate the verifier, as the latter is computationally efficient (\Cref{rem:verifier_efficiency}).

Importantly, we may \emph{not} assume that the verifier $V$ is differentiable---it is an arbitrary (efficient oracle) Turing machine---and so we cannot directly compute gradients of its decision with respect to model parameters. The challenge is to align the model with a verifier.
\Cref{alg:TL,alg:RLVF} address this challenge by (essentially) computing unbiased estimators for the Verifiability $\ver_V(\theta)$ or a surrogate (lower-bound) thereof. We formally prove these properties in \Cref{lem:grad_restated,lem:grad_rlvf}.

Our approach is inspired by Reinforcement Learning from Human Feedback \citep{ChristianoLBMLA17}, a method for aligning models with human preferences, which has recently been used to align sequence-to-sequence models \citep{Ouyang0JAWMZASR22}.
However, there are two important differences between humans and algorithmic verifiers: (1) Verifiers are efficient algorithms which may be emulated by the learner. This is unlike humans, whose preferences are costly to obtain. On the other hand, (2) verifiers make a single-bit decision at the end of an interaction, but cannot guide the prover (model) in intermediate rounds. In RL terms, this is known as the \emph{exploration problem} for sparse reward signals (e.g. \citealt{LadoszWKO22}).

The full specification of the learning model can be found in \Cref{sec:learning_model}.
We will refer to the \emph{transcript} of an interaction between a verifier and a prover (see \Cref{fig:main}), denoted by $\pi = (y,q_1,a_1,\dots,q_R,a_R)$. Let $\pi_{<s} \in \Tok^{s-1}$ denote the $s$-token long prefix of $\pi$.

\subsection{Transcript Learning}\label{sec:ver_base}
We first present an algorithm for learning Self-Proving models which relies on access to a distribution of accepting transcripts. We focus on the algorithm first, and then discuss how the learner may obtain accepting transcripts in \Cref{sec:impl_trans_generator}.
The idea is to learn a model not just of $x \mapsto y^\ast$ for a correct output $y^\ast$, but of $x \mapsto y^\ast\pi^\ast$, where $\pi^\ast$ is a transcript of an interaction in which the verifier accepted. Formally, Transcript Learning assumes access to a \emph{transcript generator}---a random variable over transcripts that faithfully represents the interaction of the verifier with some prover for a given input. 
An \emph{honest transcript generator} is one which is fully supported on transcripts accepted by the verifier. These are defined next.
\begin{definition}[Transcript generator]\label{def:trans_gen}
	Fix a verifier $V$ in a proof system of $R \in \N$ rounds. A transcript generator $\Transcript_V$ for $V$ is a randomized mapping from inputs $x \in \Seq$ to transcripts $\pi = (y,q_1,a_1,\dots,q_R,a_R) \in \Seq$. For any input $x$, $\Transcript_V(x)$ satisfies that for each $r \leq R$, the marginal of $\Transcript_V(x)$ on the $r^{\text{th}}$ query $q_{r}$ agrees with the corresponding marginal of the query generator $(V_q)_r$.\footnote{A \emph{query generator} $V_q$ corresponding to $V$ takes as input a partial interaction and samples from the distribution over next queries by $V$. Formally, for any $r \leq R$, given input $x$, output $y$, and partial interaction $(q_i,a_i)_{i=1}^{r}$, $V_q(x,y,q_1,a_1,\dots,q_{r},a_{r})$ is a random variable over $\Sigma^{L_q}$. For completeness' sake, we can say that when prompted with any sequence $z$ that does not encode an interaction, $V_q(z)$ is fully supported on a dummy sequence $\bot \cdots \bot \in \Tok^{L_q}$.}
	A transcript generator $\Transcript^\ast_V \defeq \Transcript_V$ is \emph{honest} if it is fully supported on transcripts $\pi^\ast$ for which the verifier accepts.
\end{definition}

\begin{figure}[ht]
    \centering
    \includegraphics[width=1\linewidth]{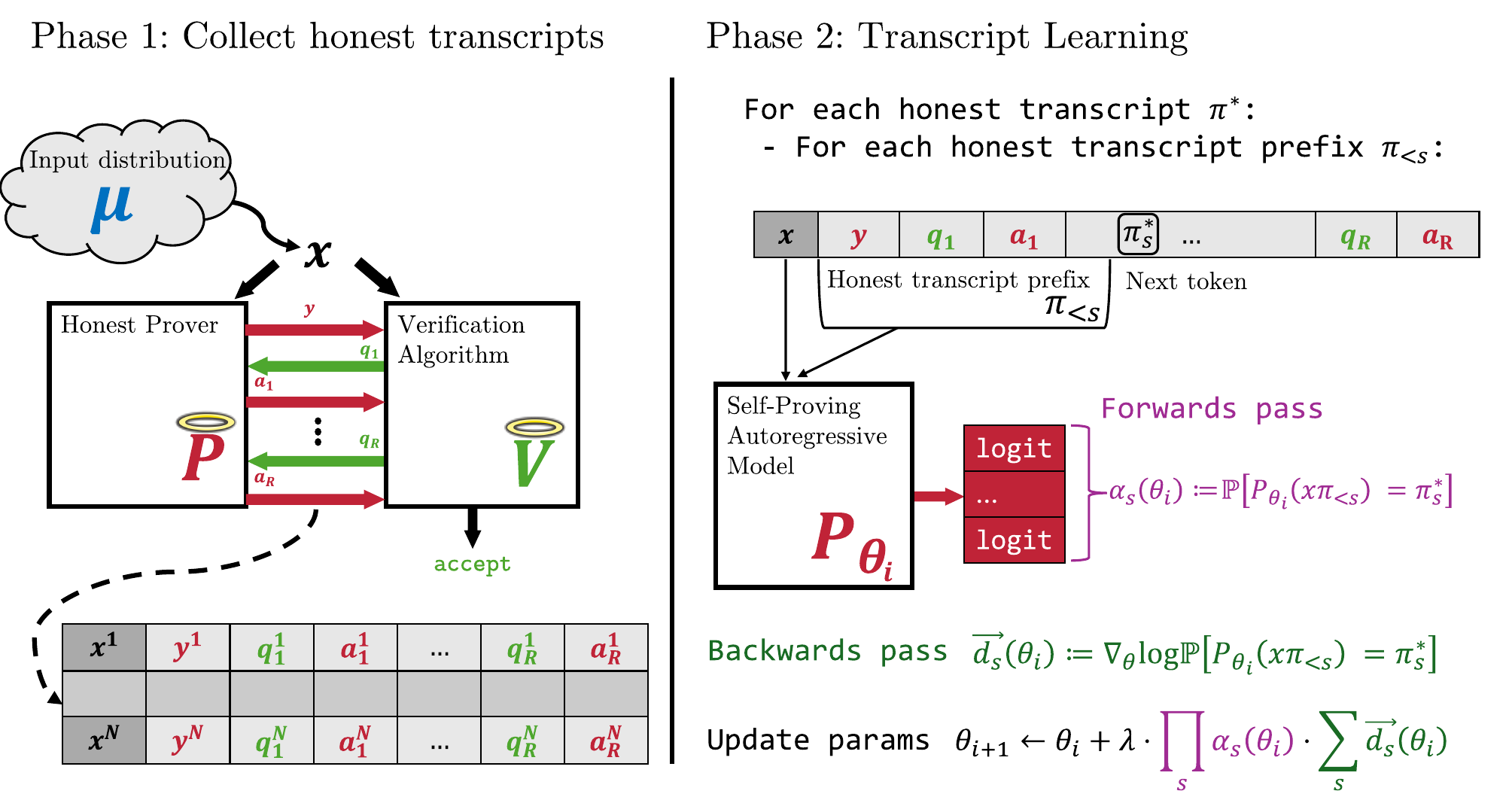}
    \caption{\captitle{Transcript Learning, visualized.} To understand \Cref{alg:TL}, consider the above visualization. In Phase 1, $N$ honest transcripts are collected by interacting an honest prover with the Verifier; these serve as samples from the transcript generator $\Transcript^\ast_V(x)$. Phase 2 runs \Cref{alg:TL}: for each transcript $\pi^\ast$ (lines 2–3) and each prefix $\pi_s$ (lines 4–6), the values $\alpha_s(\theta_i)$ and $\vec{d}_s(\theta_i)$ are computed via forward and backward passes (line 7), followed by a parameter update (line 8).}
    \label{fig:tl}
\end{figure}

Notice that for any verifier $V$, there is a one-to-one correspondence between transcript generators and (possibly randomized) provers. We intentionally chose \emph{not} to specify a prover in \Cref{def:trans_gen} to emphasize that transcripts can be ``collected'' independently of the honest prover (see completeness in \Cref{def:sys-ver}), and in fact can be collected ``in advance'' prior to learning (see \Cref{fig:tl}). As long as the generator is fully supported on honest transcripts, it can be used for Transcript Learning as depicted in \Cref{alg:TL} and \Cref{fig:tl}.

\begin{algorithm}[ht]
   \caption{Transcript Learning (TL)}
   \label{alg:TL}
   \KwHyperparameters{Learning rate $\lambda \in (0, 1)$ and number of samples $N \in \N$.}
   \KwInput{An autoregressive model family $\{P_\theta\}_{\theta \in \mathbb{R}^d}$, verifier specification (code) $V$, and sample access to an input distribution $\mu$ and an accepting transcript generator $\Transcript_V^{\ast}( \cdot)$.}
   \KwOutput{A vector of parameters $\bar{\theta} \in \mathbb{R}^d$.}
   \SetKwBlock{Begin}{begin}{end}

   Initialize $\theta_0 \defeq \vec{0}$. \\
   \For{$i = 0,\dots,N-1$}{
           Sample $x \gets \mu$ and $\pi^* = (y^*,q^*_1,a^*_1,\dots,q^*_R,a^*_R) \sim \Transcript_V^\ast(x)$. Denote $a_0 \defeq y^*$. \\
           \ForEach{Round of interaction $r = 0,\dots, R$}{
                Let $S(r)$ denote the indices of the $r^{\text{th}}$ answer $a_r$ in $\pi^\ast$, and let $\pi_{<s}$ denote the prefix of the partial transcript $(y,q^*_1,a_1^*,\dots,q^*_r)$.\\
                \For{$s \in S(r)$}{
                Compute
                \algorithmiccomment{Forwards and backwards pass} \begin{align*}
                	\alpha_s(\theta_i) &\defeq \Pr_{\sigma \gets p_{\theta_i}(x \pi_{<s})}[\sigma = \pi^*_s] \\
                	\vec{d}_s(\theta_i) &\defeq \nabla %
                 _{\theta}  \log \alpha_s(\theta_i) = \nabla_{\theta} \log \Pr_{\sigma \gets p_{\theta_i}(x \pi_{<s})}[\sigma = \pi^*_s].
                \end{align*}
				}
                 }
           Update \begin{equation*}
           		\theta_{i+1} \defeq \theta_i + \lambda \cdot  \prod_{\substack{r \in [R] \cup \{0\} \\ s \in S(r)}} \alpha_{s}(\theta_i) \cdot
           		\sum_{\substack{r \in [R] \cup \{0\} \\ s \in S(r)}}\vec{d}_s(\theta_i).
           \end{equation*}
           }
   Output $\bar{\theta} \defeq \frac{1}{N} \sum_{i \in [N]} \theta_i$. 
\end{algorithm}

TL trains a Self-Proving model by autoregressively optimizing towards generating accepting transcripts. At a high level, it works by repeatedly sampling $x \gets \mu$ and $y^\ast \pi^\ast \gets \Transcript^\ast(x)$, and updating the logits $\log p_\theta$ towards agreeing with $y^\ast \pi^\ast$ via Gradient Ascent.
While TL does not directly estimate the Verifiability gradient $\nabla_\theta \ver_V(\theta)$, we are able to show that it estimates a gradient of a lower-bounding function $A(\theta) \leq \ver_V(\theta)$. Therefore, as it ascends the gradient, it optimizes $\ver_V(\theta)$ via the surrogate. Formally, the lower-bounding function $A(\theta)$ is the agreement of the transcripts generated by the current model $P_\theta$, with the transcripts generated by the honest transcript generator. That is, $A(\theta) \defeq \Pr[\pi = \pi^*]$ where the probability is over $x \gets \mu$, $\pi_\theta \gets \Transcript^\theta_V(x)$, and $\pi^* \gets \Transcript^\ast_V(x)$.

\begin{lemma}[TL gradient estimation]\label{lem:grad_restated}
Fix an input distribution $\mu$ over $\Seq$ and a verifier $V$ with round complexity $R$ and answer length $L_a$. Fix an honest transcript generator $\Transcript_V^\ast$. 
Let $\theta$ denote the parameters of the model $P_\theta$, let $A(\theta)$ be as defined above and let the terms $S(r)$, $\alpha_s(\theta)$, and $\vec{d}_s(\theta)$ be as defined in \Cref{alg:TL}. Then,
\begin{equation*}
	\nabla A(\theta) =
	\E_{\substack{x \gets \mu \\ \pi^\ast \gets \Transcript^\ast_V}} \left[\prod_{\substack{r \in [R]\cup\{0\} \\ s \in S(r)}} \alpha_s(\theta) \cdot
           		\sum_{\substack{r \in [R] \cup \{0\} \\ s \in S(r)}}\vec{d}_s(\theta) \right].
	\end{equation*}
\end{lemma}
The proof is deferred to \Cref{apx:grad_restated}. Note that \Cref{lem:grad_restated} is true for \emph{any} model $P_\theta$. Moreover, the random vector over which the expectation is taken (in the right hand side) is precisely the direction of the update performed in \Cref{alg:TL}. In \Cref{sec:converge_TL}, we will use \Cref{lem:grad_restated} to prove convergence bounds for TL under certain conditions.

\subsubsection{Access to Accepting Transcripts}\label{sec:impl_trans_generator}
As mentioned, Transcript Learning relies on access to accepting transcripts. In this section we discuss how such access can be realized (grounded in the theory of Interactive Proofs).

\paragraph{Doubly-Efficient Interactive Proofs.}
\textit{When the honest prover $P^\ast$ is efficient (polynomial time)}, the learner (who has the code for $P^\ast$ and $V$) can execute $P^\ast$ on input $x$ to collect accepting transcripts---asuming no distribution shift at inference time. This setting is formalized by the notion of \emph{Doubly-Efficient Interactive Proofs} (DEIPs), introduced in the foundational work of ~\citet{GKR08}, who construct DEIPs for all problems computable by log-space uniform families of polynomial-size circuits with polylogarithmic depth. Their protocols ensure that the verifier runs in nearly linear time, while the prover operates in polynomial time—or more generally, in time proportional to the circuit size. Later, ~\citet{RRR16} showed that any problem computable by a Turing machine running in polynomial time and sublinear space admits a constant-round DEIP. Subsequent theoretical works (e.g., \citealt{GR18,GoldreichR18}) have constructed DEIPs for specific problems, while applied works (e.g., \citealt{ZhangLWZSXZ21,Thaler}) have improved the time and space complexity of such protocols in practice.

\paragraph{Backward Data Generation.}
Recent works on ``AI for advanced mathematical problems'' \citep{charton2021,Lyapunov} propose to reverse the generation process of problem-solution pairs as follows: rather than sampling problems and searching for solutions, one first samples solutions and then constructs the corresponding problems.  In our setting, this inspires the following approach. Suppose it is computationally hard in the worst case to start with the input $x$ and generate an accepting transcript $\pi = (y,q_1,a_1,\dots,q_R,a_R)$ for $x$.  Could we instead jointly sample $(x,\pi)$, or first sample a transcript $\pi$ and then extract the input $x$ for which $\pi$ is an accepting transcript? As long as the resulting $x$'s are distributed the same as inference time,  this would enable Transcript Learning of such instances.

We explain the idea further with a cryptographic example.  Setting up a Diffie--Hellman key-exchange scheme for security parameter $1^k$ requires producing as a public parameter a prime $p$ in a factored form, namely $p-1=\Pi{{q_i}^{\alpha_i}}$ where $q_i$ are primes. However, factoring $(p-1)$ is generally hard.  Instead, one could go ``backwards:'' first generate $(p-1)$ in factored form by choosing primes $q_i$ and exponents $\alpha_i$ \citep{kalai2003generating} and then testing if  $p=1+\Pi{{q_i}^{\alpha_i}}$ is prime. By the prime number theorem, $p$ is likely to be prime after a few attempts. In the context of our paper, one may ask a Self-Proving model to produce certified primes, training it on tuples $(x \defeq p,\pi \defeq ( q_i,\alpha_i)_i)$.

\subsection{Reinforcement Learning from Verifier Feedback (RLVF)}\label{sec:ver_amp}

As mentioned in \Cref{sec:ver_base}, Transcript Learning uses access to an honest transcript generator to estimate gradients of (a lower bound on) the Verifiability of a model $P_\theta$. 
Next we present \emph{Reinforcement Learning from Verifier Feedback} (RLVF, \Cref{alg:RLVF}), which estimates this gradient without access to a transcript generator.

\begin{algorithm}[ht]
   \caption{Reinforcement Learning from Verifier Feedback (RLVF)}
   \label{alg:RLVF}
   \KwHyperparameters{Learning rate $\lambda \in (0, 1)$ and number of samples $N \in \N$.}
   \KwInput{An autoregressive model family $\{P_\theta\}_{\theta \in \mathbb{R}^d}$, initial parameters $\theta_0 \in \mathbb{R}^d$, verifier specification (code) $V$, and sample access to an input distribution $\mu$.}
   \KwOutput{A vector of parameters $\bar{\theta} \in \mathbb{R}^d$.}
      \SetKwBlock{Begin}{begin}{end}

   \For{$i = 0,\dots,N-1$}{
           Sample $x \gets \mu$. \\
           Initialize $a_0 \defeq y \gets P_{\theta_i}(x)$. \\
           \ForEach{Round of interaction $r = 1,\dots R$}{
           		Sample the $r^{\text{th}}$ query \algorithmiccomment{Emulate the verifier} \begin{equation*}
           			q_{r} \gets V_q(x,a_0,q_1,a_1,\dots,q_{r-1},a_{r-1}).
           		\end{equation*}
           		Sample the $r^{\text{th}}$ answer \algorithmiccomment{Forwards pass}\begin{equation*}
           			a_{r} \gets P_{\theta_i}(x,a_0,q_1,a_1,\dots,q_r).
           		\end{equation*}
                Let $\tau_r \defeq (a_0,q_1,\dots,a_{r-1},q_{r})$.\\
                \For{$s \in [L_a]$}{
                Let $a_{r,s}$ denote the $s^{\text{th}}$ token in $a_r$. Compute \algorithmiccomment{Backwards pass} \begin{align*}
                	\vec{d}_s(\theta_i) &\defeq \nabla_\theta \log \Pr_{\sigma \gets p_{\theta_i}(x \tau_r)}[\sigma = a_{r,s}].
                \end{align*}
				}
				}
				\If{$V(x,y,q_1,a_1,\dots,q_R,a_R)\ \accepts$}{
				Update \begin{equation*}
           		\theta_{i+1} \defeq \theta_i + \lambda \cdot \sum_{\substack{r \in [R] \cup \{0\} \\ s \in [L_a]}}\vec{d}_s(\theta_i).
                                 \end{equation*}
           }}
   Output $\bar{\theta} \defeq \frac{1}{N} \sum_{i \in [N]} \theta_i$.
\end{algorithm}

Note that the parameters are updated (line 11) only when an accepting transcript was generated. This means that the learner can first fully generate the transcript (lines 6-7), and then take backwards passes (line 9) only if the transcript was accepted by $V$. This is useful in practice (e.g. when using neural models) as backwards passes are more computationally expensive than forwards passes.

On the other hand, this means that RLVF requires the parameter initialization $\theta_0$ to have Verifiability bounded away from 0, so that accepting transcripts are sampled with sufficient probability. Fortunately, such a Self-Proving base model can be learned using TL. This gives a learning paradigm in which a somewhat-Self-Proving base model is learned with TL (with Verifiability $\delta \gg 0$), and then ``amplified'' to a fully Self-Proving model using RLVF. This can be seen as an adaptation of the method of \citet{NairMAZA18} to the setting of Self-Proving models.

When comparing \Cref{alg:TL,alg:RLVF}, we see that the latter (RLVF) does not keep track of the probabilities $\alpha_s$. This is because, in RL terms, RLVF is an \emph{on-policy} algorithm; it generates transcripts using the current learned model, unlike TL that samples them from a distribution whose parameterization is unknown to the learner. Hence, the update step in RLVF is simpler than TL. 

We show that the update step in RLVF maximizes the Verifiability of $P_\theta$.
\begin{lemma}[RLVF gradient estimation]\label{lem:grad_rlvf}
	Fix an input distribution $\mu$ over $\Seq$ and a verifier $V$ with round complexity $R$ and answer length $L_a$. For any transcript $(x,y,q_1,\dots,a_R)$ we let $\Acc_V(x,y,q_1,\dots,a_R)$ denote the indicator random variable which equals 1 if and only if $V$ accepts the transcript.
	For any model $P_\theta$, denote by $\ver(\theta)$ the verifiability of $P_\theta$ with respect to $V$ and $\mu$ (\Cref{def:ver_mod}). Then, for any $\theta$, 
\begin{equation*}
	\nabla_\theta \ver(\theta) =
	\E_{\substack{x \gets \mu \\ y \gets P_\theta(x) \\ (q_r,a_r)_{r=1}^R}} \left[ \Acc_V(x,y,q_1,\dots,a_R) \cdot
           		\sum_{\substack{r \in [R] \cup \{0\} \\ s \in [L_a]}}\vec{d}_s(\theta) \right]
	\end{equation*}
	where $(q_r, a_r)_{r=1}^R$ are as  in lines 5-6 of \Cref{alg:RLVF}, and $\vec{d}_s(\theta)$ is as defined in line 8 therein.
\end{lemma}
Note that, because $\Acc_V(\cdot)$ is a 0-1 indicator of whether a transcript was accepted, then the right hand side of the above equation is precisely the direction of the step taken in RLVF (line 9). The proof of \Cref{lem:grad_rlvf} can be found in \Cref{apx:rlvf}.

\section{Convergence of Transcript Learning} \label{sec:converge_TL}
As an application of \Cref{lem:grad_restated}, we prove that, under certain conditions, Transcript Learning (TL, \Cref{alg:TL}) is expected to output a Self-Proving model. While the theorem relies on simplifying assumptions such as convexity and Lipschitzness, it offers clean mathematical guarantees that illuminate the core dynamics of Transcript Learning. As we discuss following the theorem statement, such conditions are common in theoretical machine learning literature, allowing us to build intuition even when the assumptions may not hold in practice.

\begin{theorem}[informal]\label{thm:TLinformal}
    Fix a verifier $V$, an input distribution $\mu$, and an autoregressive model family $\{P_\theta\}_{\theta \in \mathbb{R}^d}$. Fix an honest transcript generator $\Transcript^\ast_V$. Assume the following:
\begin{enumerate}
    \item The \emph{agreement} $A(\theta)$, informally defined as the probability that $P_\theta$ generates transcripts agreeing with $\Transcript^\ast_V$, is concave in $\theta$. Additionally, the logits of $P_\theta$ are $\Blip$-Lipschitz in $\theta$.
    \item There exist parameters $\theta^\ast$ with $||\theta^\ast|| \leq \Bnorm$ such that $P_{\theta^\ast}$ is $(1 - \eps/2)$-Self Proving.
    \item The number of tokens sent by the prover in the proof system is at most $\Cprov$.
\end{enumerate}
Then, in expectation, TL run on $O(\Cprov^2 \Bnorm^2 \Blip^2 / \eps^2)$ samples outputs a $(1-\eps)$-Self Proving model.
\end{theorem}
The full statement and proof of \Cref{thm:TLinformal} are deferred to \Cref{apx:TLmain}. Its conditions can be split into two parts.
First (item 1), convexity and Lipschitzness, which are simplifying assumptions commonly needed to prove SGD convergence. While convexity does not hold in general for DNNs, analyzing convex settings provides clean mathematical tools for establishing foundational results---an approach commonly used in ML theory, particularly for DNNs. Indeed, several works have addressed the problem of proving convergence without convexity \citep{du2019gradient,bartlett2006convexity,khaled2023better}.

Norm-boundedness (item 2), on the other hand, is a \emph{(necessary) realizability assumption}: if the architecture $\{P_\theta\}_\theta$ cannot be instantiated with parameters $\theta^\ast$, then it cannot be trained to be Self-Proving. This assumption is well-grounded for transformer architectures, as recent theoretical work has established their Turing-completeness \citep{BhattamishraPG20,DehghaniGVUK19}.

Finally (item 3), the bound $\Cprov$ on the communication complexity of the prover in the Interactive Proof system. This parameter directly affects the efficiency of TL, as reflected in the \emph{number of iterations} (and sampled transcripts): it depends on both the \emph{optimization landscape complexity} $\Bnorm^2 \Blip^2 / \eps^2$ and the \emph{communication complexity} $\Cprov^2$. Reducing communication has long been a central objective in the study of proof systems (e.g., \citealt{GoldreichH98,GoldreichVW02,RRR16}). \Cref{thm:TLinformal} formalizes how communication-efficient proof systems improve the performance of Self-Proving models.

\begin{remark}[Towards a convergence theorem for RLVF]
	RLVF can be derived by viewing Self-Proving as a reinforcement learning problem in which the agent (prover) is rewarded when the verifier accepts. Indeed, RLVF is the Policy Gradient method \citep{SuttonMSM99} for a verifier-induced reward. Convergence bounds for Policy Gradient methods are a challenging and active area of research (e.g. \citealt{AgarwalKLM21}), and so we leave the full analysis to future work.
\end{remark}

\section{Learning From Annotations}\label{apx:annot}

To minimize the length of messages exchanged in an Interactive Proof system, the honest prover is designed to send the shortest possible message to the verifier, containing only essential information.

However, when training Self-Proving model, it may be useful for it to first generate an ``annotated'' answer $\widetilde{a}$ which is then trimmed down to the actual answer $a$ to be sent to the verifier. We adapt \Cref{sec:defs,sec:learning_ver_mod} to this setting via \emph{Annotated Transcripts}. The TL and RLVF algorithms naturally extend to annotated transcripts as well. \Cref{tab:main} shows that annotations significantly improve performance of TL in practice.

Annotations can be viewed as adding Chain-of-Thought \citep{Wei0SBIXCLZ22}. As a concrete example, consider our experiments on computing the GCD. As detailed in \Cref{sec:gcd_annot}, a proof $\pi$ in this setting is the output of an iterative process---the extended Euclidean algorithm---starting from the input $x$: $x \mapsto \pi_1 \mapsto \pi_2 \mapsto \dots \mapsto \pi$. The annotation of the proof $\pi$ consists the first $T$ steps $(\pi_1,\dots, \pi_T)$ up to some fixed cutoff $T$. These are prepended to the proof and shown to the model during TL training. At inference time, the model is evaluated only on whether it generated the proof $\pi$ correctly.

We formally capture annotations by introducing a \emph{transcript annotator} and an \emph{answer extractor} incorporated into the training and inference stages, respectively.
Fix a verifier $V$ in an $R$-round proof system with question length $L_q$ and answer length $L_a$. An \emph{annotation system} with annotation length $\widetilde{L_a}$ consists of a \emph{transcript annotator} $A$, and an \emph{answer extractor} $E$.

In terms of efficiency, think of the annotator as an algorithm of the same computational resources as an honest prover in the system (see \Cref{def:sys-ver}), and the answer extractor as an extremely simple algorithm (e.g., trim a fixed amount of tokens from the annotation).

To use an annotation system the following changes need to be made:
\begin{itemize}
    \item At training time, an input $x$ and transcript $\pi$ is annotated to obtain $\widetilde{\pi} \defeq A(x,\pi)$, e.g. before the forwards backwards pass in TL (line 3 in \Cref{alg:TL}).
    \item At inference time (i.e., during interaction between $V$ and $P_\theta$), the prover keeps track of the annotated transcript, but in each round passes the model-generated (annotated) answer through the extractor $E$ before it is sent to the verifier. That is, in each round $r \in [R]$, the prover samples
    \begin{equation*}
        \widetilde{a_{r}} \gets P_{\theta}(x, y, q_1, \widetilde{a_1}, \dots, \widetilde{a_{r-1}}, q_r).
    \end{equation*}
    The prover then extracts an answer $a_r \defeq E(\widetilde{a_r})$ which is sent to the verifier.
\end{itemize}

\section{Experiments}\label{apx:experiments}

\begin{table}[t]
\caption{\textbf{Self-Proving transformers computing the GCD.} We train a 6.3M parameter GPT to compute the GCD of two integers sampled log-uniformly from $[10^4]$. Vanilla GPT correctly generates the GCD for almost all inputs, but does not prove correctness to a simple verification algorithm. GPT trained with Transcript Learning (GPT+TL) proves its answer 60.3\% of the time; adding Reinforcement Learning from Verifier Feedback (+RLVF) increases this to 78.3\%; training with Annotated Transcript Learning (GPT+ATL) gives the highest Verifiability score of 96\%.}
\label{tab:main}
\vskip 0.15in
\begin{center}
\begin{small}
\begin{sc}
\begin{tabular}{lccr}
\toprule
Learning method & Correctness & Verifiability \\
\midrule
GPT (baseline)        & 99.8\% & -     \\
GPT+TL                & 98.8\% & 60.3\%  \\
GPT+TL+RLVF           & 98.9\% & 78.3\%  \\
GPT+ATL      & 98.6\% & 96.0\%  \\
\bottomrule
\end{tabular}
\end{sc}
\end{small}
\end{center}
\vskip -0.1in
\end{table}

We describe our experimental setup, and present ablation studies that shed additional light on the effect of \emph{annotation} and \emph{representation} on Verifiability.

\subsection{Setup: Training transformers to predict the GCD}
\citet{Charton24} empirically studies the power and limitations of learning GCDs with transformers. We follow their setup and two conclusions on settings that make for faster learning: Training from the log-uniform distribution, and choosing a base of representation with many prime factors.

We fix a base of representation $B = 210$ and use $\seq{x}$ to denote an integer $x$ encoded as a $B$-ary string.\footnote{$B=210$ is chosen following \citet{Charton24} to be an integer with many prime factors.} For sequences of integers, we write $\seq{(x_1x_2)}$ to denote the concatenation of $\seq{x_1}$ with $\seq{x_2}$, delimited by a special token. The vocabulary size needed for this representation is $|\Sigma| \approx 210$.

We choose the input distribution $\mu$ to be the log-uniform distribution on $[10^4]$, and train the transformer on sequences of the form $\seq{(x_1 x_2 y)}$, where $x_1,x_2 \sim \mu$ and $y=GCD(x_1,x_2)$.
This is a scaling-down of \citet{Charton24}, to allow single GPU training of Self-Proving transformers. In all of our experiments, we use a GPT model \citep{VaswaniSPUJGKP17} with 6.3M parameters trained on a dataset of 1024K samples in batches of 1024. Full details are deferred to \cref{apx:experiments-details}.

\paragraph{Proving correctness of GCD.}
Following \citet{Charton24} as a baseline, we find that transformers can correctly compute the GCD with over $99\%$ probability over $(x_1,x_2) \gets \mu$. To what extent can they \emph{prove} their answer? To answer this question, we first devise a natural proof system based on Bézout's theorem. Its specification and formal guarantees are deferred to \cref{app:verifying_the_GCD}. We denote its verification algorithm by $V$, and highlight some important features of the experimental setup:
\begin{itemize}
	\item The proof system consists of one round ($R=1$). The verifier makes no query, and simply receives a proof $\seq{\pi}$ from the prover.
	\item \emph{Completeness:} For any $x_1,x_2,y \in [10^4]$ such that $y = GCD(x_1,x_2)$, there exists a proof $\seq{\pi}$ such that $V(\seq{x_1x_2 y \pi})$ accepts. As detailed in \Cref{app:verifying_the_GCD}, the proof $\pi$ consists of a pair of integers who are \emph{Bézout coefficients} for $x_1,x_2$.
	\item \emph{Soundness:} If $y \neq GCD(x_1,x_2)$, then $V(\seq{x_1 x_2 y \pi})$ rejects\footnote{With probability 1, i.e., $s=0$ in \cref{def:sys-ver}.} for any alleged proof $\seq{\pi} \in \Seq$.
\end{itemize}

To measure Verifiability, we train a Self-Proving transformer using Transcript Learning on sequences $\seq{(x_1 x_2 y \pi)}$ and estimate for how many inputs $x_1, x_2 \gets \mu$ does the model generate $\emph{both}$ the correct GCD $\seq{y}$ and a valid proof $\seq{\pi}$. We test on 1000 pairs of integers $x_1',x_2' \gets \mu$ held-out of the training set, prompting the model with $\seq{(x_1'x_2')}$ to obtain $\seq{(y' \pi')}$, and testing whether $V(\seq{x_1'x_2' y' \pi'})$ accepts.

\Cref{tab:main} shows our main experimental result, which has the following key takeaways:
\begin{enumerate}
    \item Transcript Learning (TL) for 100K iterations ($\approx\!$100M samples) results in a Self-Proving transformer that correctly proves 60.3\% of its answers.
    \item A base Self-Proving Model with fairly low Verifiability of 40\% can be improved to 79.3\% via Reinforcement Learning from Verifier Feedback (RLVF). Although it does not rely on honest transcripts, RLVF trains slowly: this nearly-twofold improvement took four million iterations.
    \item Most efficient is Annotated Transcript Learning, which yielded a model with 96\% Verifiability in 100K iterations. We further investigate their effect next.
\end{enumerate}

\subsection{Models generalize beyond annotations}\label{sec:gcd_annot}

\begin{figure}[ht]
    \centering
    \includegraphics[width=1\columnwidth]{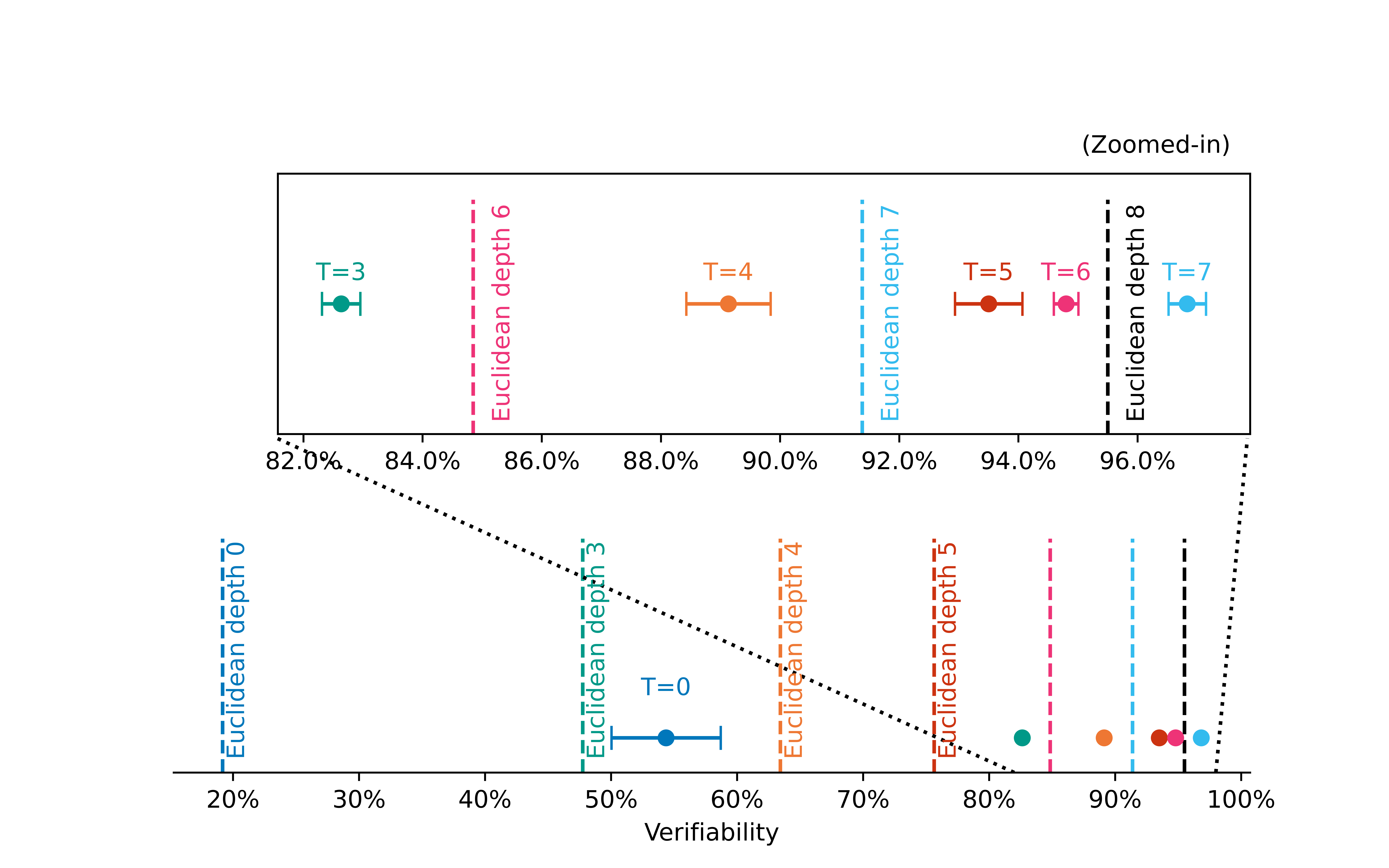}
    
    \caption{\textbf{Verifiability with increasing amounts of annotation}. $T$ is the number of steps added in Annotated Transcript Learning. Dashed lines indicate \emph{Euclidean depth}, that bound the Verifiability of models that prove \emph{only} for integers up to a certain number of steps. Each $T$ was run with three seeds, with mean $\pm$ standard error depicted. The upper graph provides a zoomed-in view of the 82\% to 98\% range from the lower graph, which spans a broader scale from 20\% to 100\%.}
    \label{fig:annot}
\end{figure}

\begin{figure}[htb]
    \centering
   \includegraphics[width=1\columnwidth]{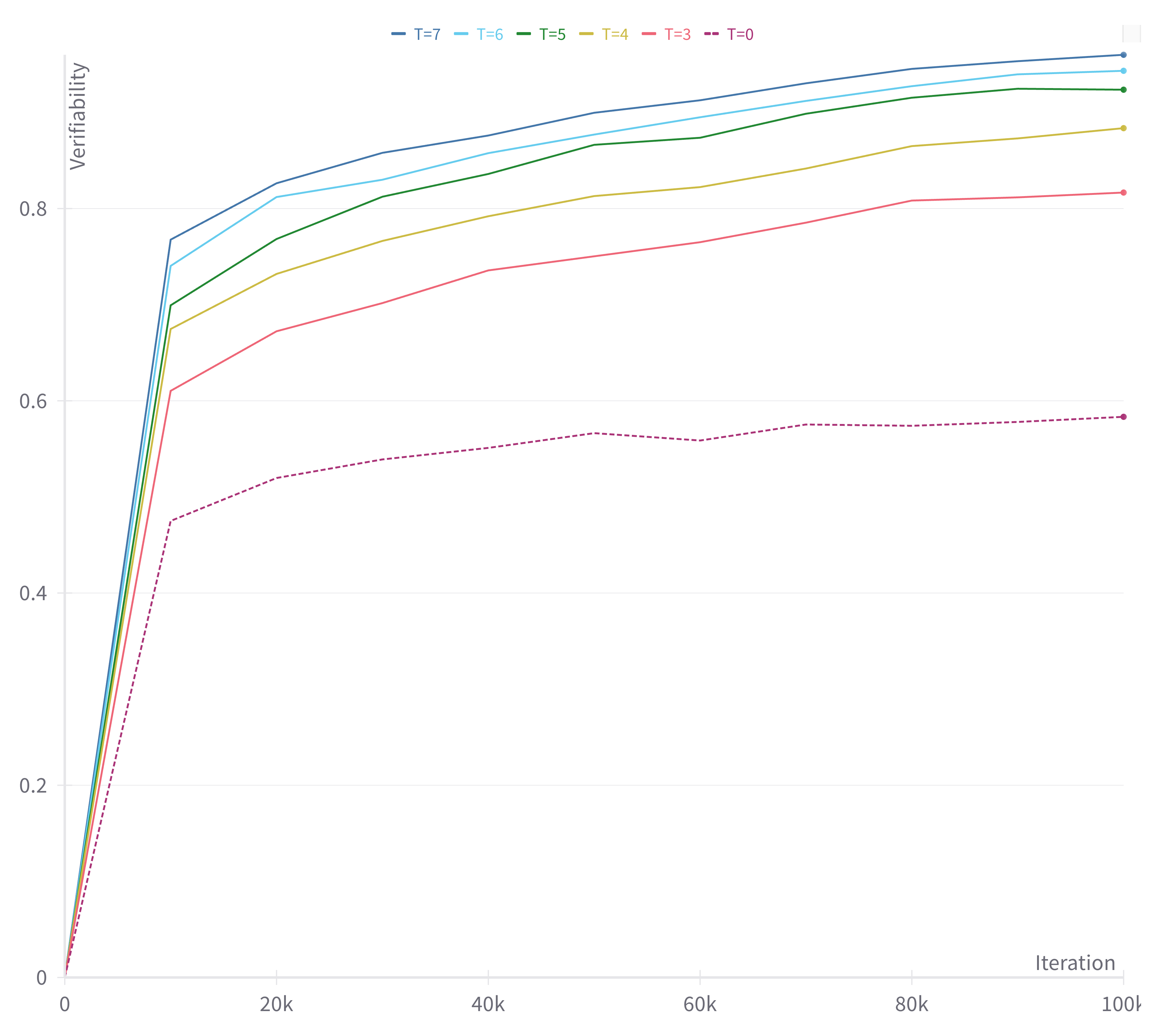}
    \caption{\textbf{Annotated TL Verifiability as a function of the number of samples $N$}. Each iteration (X axis) is a batch of 1024 samples from a dataset of $\approx$10M sequences. Every 10k iterations, Verifiability was evaluated on a held-out dataset of 1k inputs. $T$ is the number of steps in Annotated Transcript Learning (\Cref{fig:annot}), and $T=0$ is non-annotated Transcript Learning. Each $T$ was run with three seeds, with mean depicted by the curve and standard error by the shaded area.}\label{fig:annot_curves}
\end{figure}

The proof $\seq{\pi}$ is annotated by including intermediate steps in its computation. Details are deferred to \Cref{app:verifying_the_GCD};
roughly speaking, we observe that the proof $\seq{\pi}$ for input $\seq{(a,b)}$ is obtained as the last element in a sequence $\seq{a,b,\pi_1,\pi_2,\dots}$ computed by the Euclidean algorithm.
We annotate the proof $\seq{\pi}$ by prepending to it the sequence of \emph{Euclidean steps} $\seq{(\pi_1, \dots, \pi_T)}$ up to some fixed cutoff $T$.

\Cref{fig:annot} shows how $T$ affects the Verifiability of the learned model. As suggested by \citet{LeeSLLP23}, training the model on more intermediate steps results in better performance; in our case, increasing the number of intermediate steps $T$ yields better Self-Proving models.
One might suspect that models only learn to execute the Euclidean algorithm in-context. %
To rule out this hypothesis, we derive an upper bound on the possible efficacy of such limited models. This bound is based on the \emph{Euclidean depth} of integers $(x_1,x_2)$, which we define as the number of intermediate steps that the Euclidean algorithm makes before terminating on input $(x_1,x_2)$. Indeed, a model that only learns to compute (in-context) the arithmetic of the Euclidean algorithm would only be able to prove the correctness of inputs $(x_1,x_2)$ whose depth does not exceed the annotation cutoff $T$.

\Cref{fig:annot} tells a different story: For each cutoff $T$, we estimate the probability that integers $x_1,x_2 \gets \mu$ have Euclidean depth at most $T$ on $10^5$ sampled pairs. Larger annotation cutoff $T$ increases Verifiability, but all models exceed their corresponding Euclidean depth bound.

\subsection{Base of representation}
\begin{figure}[ht]
   \centering
   \includegraphics[width=\columnwidth]{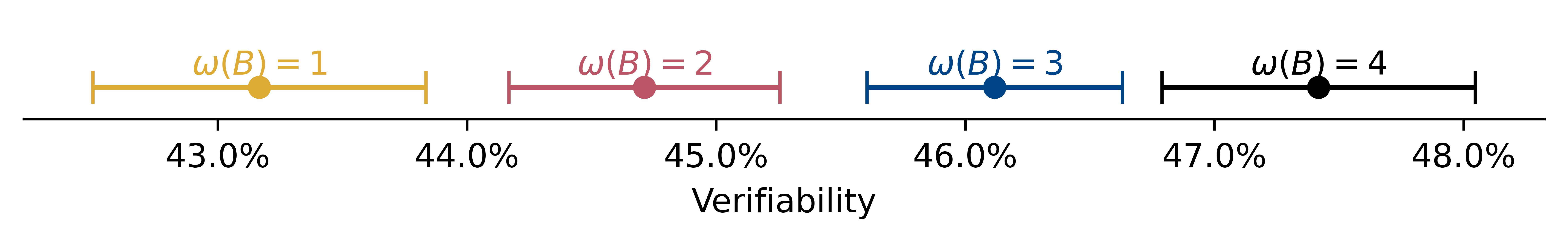}
   \caption{\textbf{The number of prime divisors of a base $\omega(B)$ determines Verifiability.} For each $o \in [4]$, we sampled 17 bases $B \in \{2, \dots, 1386 \}$ such that $\omega(B)=o$. A Self-Proving transformer was trained via Transcript Learning for twenty epochs on an identical dataset of 1024K samples encoded in base $B$. For each $\omega(B)$ we depict the mean $\pm$ standard error.}
   \label{fig:bases}
\end{figure}
As mentioned previously, \citet{Charton24} concludes that, for a given base of representation $B$, transformers correctly compute the GCD of integers $x_1,x_2$ that are products of primes dividing $B$. Simply put, choosing a base $B$ with many different prime factors yields models with better correctness (accuracy), which suggests why base $B = 210 = 2 \cdot 3 \cdot 5 \cdot 7$ yielded the best results. 
To test if $B$'s factorization has a similar effect on Verifiability, we train transformers on 68 bases varying the number of prime divisors from $\omega(B)=1$ (i.e., $B$ is a prime power) to $\omega(B)=4$. \Cref{fig:bases} shows that $\omega(B)$ correlates not just with correctness \citep{Charton24}, but also with Verifiability. Although the finding is statistically significant (no overlapping error margins), the difference is by a few percentages; we attribute this to the smaller (10\%) number of samples on which models were trained, relative to other experiments.

\begin{figure}[htb]
    \centering
   \includegraphics[width=1\columnwidth]{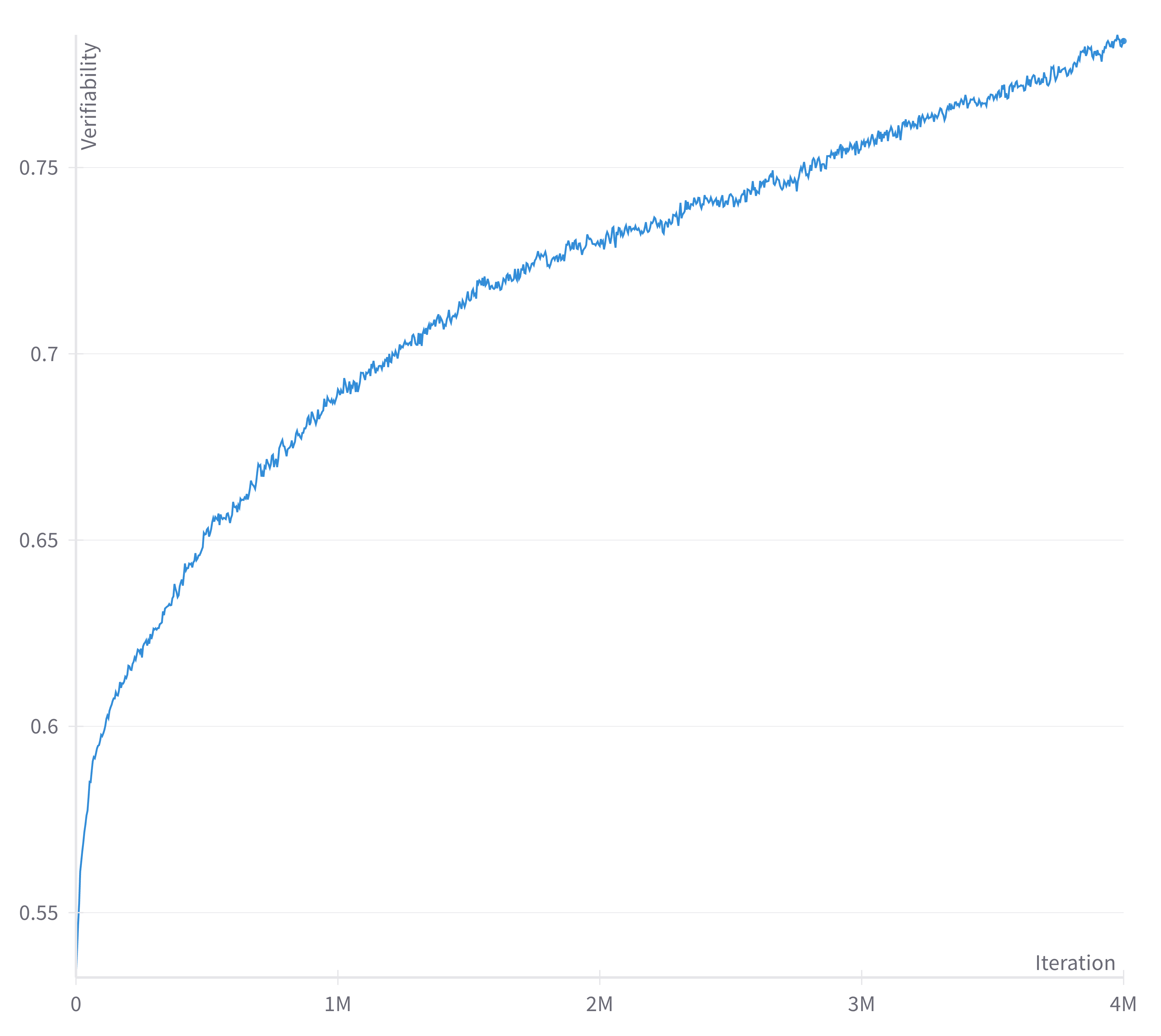}
    \caption{\textbf{RLVF Verifiability as a function of the number of samples $N$}. Starting from a base model with Verifiability 48\% (obtained via Transcript Learning), in each iteration a batch of 2048 inputs are sampled; the model generates a proof for each; the Verifier is used to check which proofs are accepted; then, the model parameters are updated accordingly (see \Cref{alg:RLVF}). Verifiability was evaluated on a held-out dataset of 1k inputs.}\label{fig:rlvf_curve}
\end{figure}

\section{Conclusion}\label{sec:conclusions}
Trust between a learned model and its user is fundamental. Interactive Proofs \citep{GoldwasserMR85} provide a general framework for establishing trust via verification algorithms. This work shows that models can be trained to formally prove their outputs within such systems---we call these \emph{Self-Proving} models. Self-Proving models connect the theory of Interactive Proofs with the goal of Trustworthy ML: they offer formal \emph{worst-case soundness guarantees}; enabling users to be confident when their models generate correct answers---and detect incorrect answers with high probability.

We support our definition with two general-purpose training methods: Transcript Learning (TL) and Reinforcement Learning from Verifier Feedback (RLVF), whose analyses draw on learning theory, RL, and computational complexity. This work can be extended in several directions: finding conditions for the convergence of RLVF, improving sample complexity bounds for TL, or designing altogether different learning algorithms (e.g., by taking advantage of properties of the verifier).

\paragraph{Limitations.}
In our current learning methods, each individual ground-truth capability requires training a separate Self-Proving model. A natural generalization of this approach is to adapt our definition and methods to deal with a single \emph{generalist} Self-Proving model that proves its correctness to multiple verifiers of different ground-truths.

Moreover, the second strategy presented in \Cref{sec:impl_trans_generator}---namely, generating accepting transcripts first and then constructing matching inputs—has an inherent limitation: the resulting training distribution is  biased by the design of the generator. As a consequence, models may end up learning  patterns of the construction process, rather than acquiring generalizable reasoning capabilities. This highlights the importance of using sufficiently diverse generators, and of evaluating model performance on out-of-distribution inputs.

\begin{ack}
We are grateful to Micah Carroll, Mark Rofin, Avishay Tal, Mojtaba Yaghoobzadeh and anonymous reviewers for their helpful comments. This research was supported by DARPA-TA1 under grant no. HR001119S0076, and by the Simons Collaboration on the Theory of Algorithmic Fairness.
\end{ack}

\bibliography{ref}
\bibliographystyle{plainnat}

\appendix

\section{Formal Proofs}
\Cref{sec:learning_model} formally specifies the setup in which our results reside. We then prove \Cref{lem:grad_restated} in \Cref{apx:grad_restated}, \Cref{lem:grad_rlvf} in \Cref{apx:rlvf}, and \Cref{thm:TLinformal} in \Cref{apx:TLmain}.
\subsection{Specification of the Learning Model} \label{sec:learning_model}
In this section, we fully specify the theoretical framework in which our results reside. We define a \emph{learner} as an algorithm $\Lambda$ with access to a family of autoregressive models $\{P_\theta\}_\theta$ and samples from the input distribution $x \gets \mu$. In our setting of Self-Proving models (and in accordance with the Interactive Proofs literature), we give the learner the full specification of the verifier $V$. More formally,
\begin{definition}[Self-Proving model learner]\label{def:learner}
	A \emph{(Self-Proving model) learner} is a probabilistic oracle Turing Machine $\Lambda$ with the following access:
	\begin{itemize}
		\item A family of \emph{autoregressive models} $\{ P_\theta \}_{\theta \in \mathbb{R}^d}$ where $d \in \N$ is the number of parameters in the family. For each $\theta$ and $z \in \Seq$, the random variable $P_\theta(z)$ is determined by  the logits $\log p_\theta(z) \in \mathbb{R}^{|\Tok|}$. For any $z \in \Seq$ and $\sigma \in \Tok$, the learner $\Lambda$ can compute the gradient of the $\sigma^{\text{th}}$ logit, that is, $\nabla_\theta \log \Pr_{\sigma' \gets p_\theta(z)}[\sigma = \sigma']$. In particular, $\log \Pr_{\sigma' \gets p_\theta(z)}[\sigma = \sigma']$ is always differentiable in $\theta$.
		
		\item Sample access to the \emph{input distribution} $\mu$. That is, $\Lambda$ can sample $x \gets \mu$.
		\item The full specification of the verifier $V$, i.e., the ability to emulate the verification algorithm $V$. More specifically, $\Lambda$ is able to compute $V$'s decision after any given interaction; that is, given input $x$, output $y$, and a sequence of queries and answers $(q_i,a_i)_{i=1}^R$, the learner $\Lambda$ can compute the decision of $V$ after this interaction.
	\end{itemize}
\end{definition}

\subsection{\texorpdfstring{Proof of \Cref{lem:grad_restated}}{Proof of Theorem 4.3}}\label{apx:grad_restated}

We let $\Transcript^\theta_V$ denote the transcript generator induced by the model $P_\theta$ when interacting with $V$: for each $x$, $\Transcript^\theta_V(x)$ is the distribution over transcripts of interactions between $V$ and $P_\theta$ on input $x$. We stress that $\pi^* \sim \Transcript^\ast_V(x)$ and $\pi \sim \Transcript^\theta_V(x)$ are transcripts produced when interacting {\em with the same verifier queries}; we can think of the verifier as simultaneously interacting with the honest prover and with the model $P_\theta$.\footnote{The way it is presented in the algorithm, first the verifier is called by $\Transcript^\ast_V$ and outputs queries $(q^*_1, \dots q^*_R)$, and then the model is prompted with the verifier queries one a time. This maintains soundness, since a proof system is sound as long as the prover does not know the verifier's queries in advance.} In what follows, we use $\pi^* \gets \Transcript^\ast_V(x)$ and $\pi \gets \Transcript^\theta_V(x)$ to denote two transcripts that share the same queries. That is, taking
$\pi^\ast = (y^\ast,q_1^\ast,a_1^\ast,\dots,q_R^\ast,a_R^\ast)$ to denote an accepting transcript sampled from $\Transcript^\ast_V(x)$, and $\pi = (y,q^*_1,a_1,\dots,q^*_R,a_R)$ to denote a random transcript sampled from $\Transcript_V^\theta(x)$, we say that $\pi$ and $\pi^*$ {\em agree} if they agree on the prover answers, namely if:
\[
(y,a_1,\dots,a_R) = (y^\ast,a_1^\ast,\dots,a_R^\ast).
\]
This definition implicitly uses the independence of the verifier and model's randomness. We now prove that TL correctly estimates the gradient of $A(\theta)$ in its update step.

\begin{proof}[Proof of \Cref{lem:grad_restated}]
Throughout this proof, expectations and probabilities will be over the same distributions as in the lemma statement. First, we use the law of total probability together with the autoregressive property of $P_\theta$ (\Cref{sec:learning_ver_mod}) to switch from probabilities on transcripts, to products of next-token probabilities.
Formally, consider a fixed input $x$, an honest transcript $\pi^\ast = (y^\ast,q_1^\ast,a_1^\ast,\dots,q_R^\ast,a_R^\ast)$, and denote a random transcript sampled from $\Transcript_V^\theta(x)$ when using the same verifier queries by $\pi = (y,q_1^\ast,a_1,\dots,q_R^\ast,a_R)$. For any $r \in [R]$ denote the random variable $\Transcript^{\theta,<r}_V \defeq \Transcript^\theta_V(yq^\ast_1a_1\cdots a_{r-1}q^\ast_r)$. Then,
\begin{align}
\Pr_{\pi}\left[  \pi=\pi^* \right] &= 
\Pr_{\pi}[(y,a_1,\dots,a_R) = (y^\ast,a_1^\ast,\dots,a_R^\ast)]  \label{eq:independence}\\
&= \Pr_{y \gets P_\theta(x)}[y = y^\ast] \cdot  \prod_{\substack{r \in [R]}}  \Pr_{a \gets \Transcript^{\theta,<r}_V}[a = a_r^\ast]  \nonumber \\
&= \Pr_{y \gets P_\theta(x)}[y = y^\ast] \cdot \prod_{\substack{r \in [R] \\ s \in S(r)}}  \Pr_{\sigma \gets p_\theta(\pi^\ast_{<s})}[\sigma = \pi^\ast_s]  \label{eq:bemet_ar} \\
&= \prod_{\substack{r \in [R]\cup\{0\} \\ s \in S(r)}} \alpha_s(\theta),
\label{eq:def_of_AR}\end{align}
where, as noted above, \Cref{eq:independence} uses the independence of the verifier and model's randomness, \Cref{eq:bemet_ar} uses the autoregressive property of $P_\theta$ (\Cref{def:learner}), and \Cref{eq:def_of_AR} is by definition of $\alpha_s$ and of $a_0$.
Next, a basic calculus identity gives
\begin{equation}\label{eq:calculus}
\nabla_{\theta} \left(\Pr_{\pi}\left[\pi=\pi^*\right] \right) = \Pr_{\pi}\left[  \pi=\pi^*\right]  \cdot \nabla_{\theta} \log \left( 
\Pr_{\pi}\left[ \pi=\pi^*\right] \right). 
\end{equation}
This implicitly assumes that $\Pr_{\pi}\left[\pi=\pi^*\right]$ is differentiable in $\theta$; indeed, this follows from \cref{def:learner}, where the logits of the model were assumed to by differentiable. 
    Let us focus on the rightmost factor. By \Cref{eq:def_of_AR},
\begin{align}\label{eq:TLgrad}
    \nabla_{\theta} \log \left(\Pr_{\pi}\left[ \pi = \pi^* \right] \right) & = 
   \nabla_{\theta} \log  \left( \prod_{\substack{r \in[R]\cup\{0\}  \\ s \in S(r)}} \alpha_s(\theta) \right) 
   	 = \sum_{\substack{r \in [R]\cup\{0\} \\ s \in S(r)}} \nabla_\theta \log \alpha_s(\theta) 
   	  =\sum_{\substack{r \in [R] \cup \{0\} \\ s \in S(r)}}\vec{d}_s(\theta) 
\end{align}
where the last equality is by definition of $\vec{d}_s(\theta)$.
Combining \Cref{eq:def_of_AR} and \Cref{eq:calculus} gives 
\[
\nabla_{\theta} \left(\Pr_{\pi}\left[\pi=\pi^*\right] \right) = \prod_{\substack{r \in [R]\cup\{0\} \\ s \in S(r)}} \alpha_s(\theta) \cdot \sum_{\substack{r \in [R] \cup \{0\} \\ s \in S(r)}}\vec{d}_s(\theta).
\]
By the law of total probability and the linearity of the gradient, 
    \begin{equation*}
\E_{x,\pi^*} \left[
\nabla_{\theta} \left(\Pr_{\pi}\left[\pi=\pi^*\right] \right) \right]=
\nabla_{\theta} \left( \E_{x,\pi^*} \left[ \Pr_{\pi}\left[\pi = \pi^* \right] \right] \right)= 
\nabla_{\theta}  \left(\Pr_{x, \pi^*, \pi}\left[\pi = \pi^* \right] \right) = \nabla_\theta A(\theta).
    \end{equation*}
which concludes the proof.
\end{proof}

\subsection{\texorpdfstring{Proof of \Cref{lem:grad_rlvf}}{Proof of Lemma 3.3}}\label{apx:rlvf}

Recall the \emph{transcript generator of $P_\theta$}, denoted by $\Transcript^\theta_V$ (see \Cref{lem:grad_restated}). By the definitions of Verifiability in \Cref{def:ver_mod} and $V(x,y,q_1,\dots,a_R)$ in the lemma statement,
	\begin{align}
		\ver(\theta) & \defeq \Pr_{\substack{x \gets \mu \\ y \gets P_\theta(x)}}\left[
		\vp{V}{P_\theta}(x,y)\  \accepts \right] \nonumber\\
		 &= 
		\E_{\substack{x \gets \mu \\ y \gets P_\theta(x) \\ (q_r,a_r)_{r=1}^R}}\left[ \Acc_V(x,y,q_1,\dots,a_R) \right] \nonumber\\ 
		&= \E_{x \gets \mu}\left[ \Pr_{\pi \gets \Transcript^\theta_V(x)}[ \Acc_V(x,\pi)\ ] \right] \label{eq:rlvf_1}
	\end{align}
	Now, for every input $x$, let $\Pi^\ast(x) \subset \Seq$ denote the set of accepting transcripts:
	\begin{equation*}
		\Pi^\ast(x) \defeq \left\lbrace \pi^\ast \in \Seq : \Acc_V({x,\pi^\ast}) =1 \right\rbrace.
	\end{equation*}
	We can assume that $\Pi^\ast(x)$ has finite cardinality, since $V$'s running time is bounded and hence the number of different transcripts that it can read (and accept) is finite. 
 For any fixed input $x$, we can express its acceptance probability by the finite sum:
	\begin{equation}\label{eq:rlvf_2}
		\Pr_{\pi \gets \Transcript^\theta_V(x)}[ \Acc_V(x,\pi)] = \sum_{\pi^\ast \in \Pi^\ast(x)} \Pr_{\pi \gets \Transcript^\theta_V(x)}[\pi = \pi^\ast].
	\end{equation}
	We will use Equations \eqref{eq:independence} through \eqref{eq:TLgrad} from the proof of \Cref{lem:grad_restated}. Up to a change in index notation, these show that, for any $\pi^\ast$,
	\begin{equation*}
		\nabla_\theta \Pr_{\pi \gets \Transcript^\theta(x)}[\pi = \pi^\ast] = \Pr_{\pi \gets \Transcript^\theta(x)}[\pi = \pi^\ast] \cdot \sum_{\substack{r \in R \cup \{0\} \\ s \in [L_a]}} \nabla_\theta \vec{d}_s(\theta).
	\end{equation*}

Combining \Cref{eq:rlvf_1,eq:rlvf_2}, by linearity of expectation we have that
	\begin{align*}
		\nabla_\theta \ver(\theta) &= \E_{x \gets \mu} \left[\sum_{\pi^\ast \in \Pi^\ast(x)} \nabla_\theta \Pr_{\pi \gets \Transcript^\theta(x)}[\pi = \pi^\ast] \right] \\
		&= \E_{x \gets \mu} \left[ \sum_{\pi^\ast \in \Pi^\ast(x)} \Pr_{\pi \gets \Transcript^\theta(x)}[\pi = \pi^\ast] \cdot \sum_{\substack{r \in R \cup \{0\} \\ s \in [L_a]}} \nabla_\theta \vec{d}_s(\theta) \right] \\
		&= \E_{x \gets \mu} \left[ \E_{\pi \gets \Transcript^\theta(x)}\left[\Acc_V(x,\pi) \cdot \sum_{\substack{r \in R \cup \{0\} \\ s \in [L_a]}} \nabla_\theta \vec{d}_s(\theta) \right]\right] \\
		&= \E_{\substack{x \gets \mu \\ \pi \gets \Transcript^\theta(x)}} \left[ \Acc_V(x,\pi) \cdot \sum_{\substack{r \in R \cup \{0\} \\ s \in [L_a]}} \nabla_\theta \vec{d}_s(\theta) \right] \\
		&= \E_{\substack{x \gets \mu \\ y \gets P_\theta(x) \\ (q_r,a_r)_{r=1}^R}} \left[ \Acc_V(x,y,q_1,\dots,a_R) \cdot \sum_{\substack{r \in R \cup \{0\} \\ s \in [L_a]}} \nabla_\theta \vec{d}_s(\theta) \right],
	\end{align*}
	where in the last equality, the probability is over $(q_r,a_r)$ sampled as in \Cref{alg:RLVF}, and it follows from the definition of the transcript generator $\Transcript^\theta(x)$.
\hfill\qedsymbol

\subsection{\texorpdfstring{Proof of \Cref{thm:TLinformal}}{Proof of Transcript Learning convergence}}\label{apx:TLmain}

We first restate \Cref{thm:TLinformal} in full formality.
\begin{theorem}\label{thm:TL}
	Fix a verifier $V$, an input distribution $\mu$, an autoregressive model family $\{P_\theta\}_{\theta \in \mathbb{R}^d}$, and a norm $||\cdot||$ on $\mathbb{R}^d$. Fix an honest transcript generator $\Transcript^\ast_V$, and assume that the \emph{agreement function}
	$\agree(\theta) \defeq  \Pr\left[\pi = \pi^* \right]$
	is concave in $\theta$, where the verifier queries are the same in $\pi^* $ and $\pi$, and the probability is over $x \gets \mu$, $\pi_\theta \gets \Transcript^\theta_V(x)$, and $\pi^* \gets \Transcript^\ast_V(x)$.
	For any $\eps > 0$, let $\Bnorm$, $\Blip$ and $\Cprov$ be upper-bounds such that the following conditions hold:
	\begin{itemize}
		\item There exists $\theta^\ast \in \R^d$ with $||\theta^\ast|| < \Bnorm$ such that $A(\theta^\ast) \geq 1 - \eps/2$.
		\item For all $\theta$, the logits of $P_\theta$ are $B_{\mathrm{Lip}}$-Lipschitz in $\theta$. That is, $\sup_{\substack{\theta \in \R^d \\ z \in \Seq}} ||\nabla_\theta \log p_\theta(z)|| \leq \Blip$.
		
		\item In the proof system defined by $V$, the total number of tokens (over all rounds) is at most $\Cprov$.
	\end{itemize}
	Denote by $\bar{\theta}$ the output of TL running for $N \geq \left( 4 \cdot \Cprov^2 \cdot \Bnorm^2 \cdot \Blip^2  \right) / \epsilon^2$ iterations and learning rate $\lambda = \Bnorm / \Cprov \Blip \sqrt{N}$. Then the expected Verifiability (over the randomness of the samples collected by TL) of $\bar{\theta}$ is at least $1 - \eps$. That is,
	$\E_{\bar{\theta}}[\ver_{V,\mu}(\bar{\theta)}] \geq 1 - \eps$.
\end{theorem}

The proof of \Cref{thm:TL} goes by reduction to Stochastic Gradient Descent (SGD). \Cref{lem:grad_restated} showed that the learner can use its only available tools---sampling honest transcripts, emulating the verifier, and differentiating the logits---to optimize the agreement $A(\theta)$. Since $A(\theta)$ lower bounds the Verifiability of $P_\theta$, the former can be used as a surrogate for the latter.

For convenience of the reader, we first provide a description of Stochastic Gradient Ascent (equiv. to SGD) and quote a theorem on its convergence. We adapt the presentation in \cite{SSS14}, noting that they present Stochastic Gradient Descent in its more general form for non-differentiable unbounded functions. The familiar reader may skip directly to the proof in \Cref{apx:actual-proof-of-TL}.

\subsubsection{Preliminaries on Stochastic Gradient Ascent}
Stochastic Gradient Ascent (SGA) is a fundamental technique in concave optimization. Given a concave function $f \colon \R^d \to [0,1]$, SGA starts at $w_0 = \vec{0} \in \R^d$ and tries to maximize $f(w)$ by taking a series of ``steps.'' Than directly differentiating $f$, SGA instead relies on an estimation $\nabla f(w)$: in each iteration, SGA takes a step in a direction that estimates $\nabla f(w)$.

\begin{definition}[Gradient estimator] \label{def:grad_estimator}
Fix a differentiable function $f \colon \R^d \to \R$ for some $d$. A \emph{gradient estimator} for $f$ is a randomized mapping $D_f \colon \R^d \to \R^d$ whose expectation is the gradient of $f$. That is, for all $w \in \R^d$,
\begin{equation*}
    \E_{v \gets D_f(w)}[v] = \nabla f(w).
\end{equation*}
Note that this is an equality between $d$-dimensional vectors.
\end{definition}

\begin{algorithm}[ht]
   \caption{Stochastic Gradient Ascent}
   \label{alg:sga}
   \KwHyperparameters{Learning rate $\lambda > 0$ and number of iterations $N \in \N$.}
   \KwInput{A function $f \colon \R^d \to \R$ to maximize and a gradient estimator $D_f$ for $f$.}
   \KwOutput{A vector $\bar{w} \in \mathbb{R}^d$.}
   \SetKwBlock{Begin}{begin}{end}

   Initialize $w_0 \defeq \vec{0} \in \R^d$. \\ 
   \For{$i = 1,\dots,N-1$}{
           Sample $v_i \gets D_f(w_{i-1})$. \\
           Update $w_i \defeq w_{i-1} + \lambda \cdot v_i$. \\
           }
    Output $\bar{w} \defeq \frac{1}{N} \sum_{i \in [N]} w_i$.
\end{algorithm}

Theorem 14.8 in \cite{SSS14} implies the following fact.
\begin{fact}\label{fact:sga}
	Fix a concave $f \colon \R^d \to [0,1]$, a norm $||\cdot||$ on $\R^d$, and upper-bounds $\Bnorm, \Blip > 0$. Let
	\begin{equation*}
		w^\ast \in \argmax_{w:||w|| < \Bnorm}f(w),
	\end{equation*}
	and let $\bar{w}$ denote the output of \Cref{alg:sga} run for $N$ iterations with learning rate
	\begin{equation*}
		\lambda = \frac{\Bnorm}{\Blip \sqrt{N}}.
	\end{equation*}
	If at every iteration it holds that $||v_i|| < \Blip$, then
	\begin{equation*}
		\E_{\bar{w}}\left[ f(\bar{w}) \right] \geq f(w^\ast) - \frac{\Bnorm \cdot \Blip}{\sqrt{N}}.
	\end{equation*}
\end{fact}

\subsubsection{Learning with Stochastic Gradient Ascent/Descent}
\Cref{fact:sga} captures the general case of using SGA for maximization of concave problems. It is more common for the literature to discuss the equivalent setting of Stochastic Gradient Descent (SGD) for minimization of convex problems. Specifically, a common application of SGD is for the task of {\em Risk Minimization}: given a loss function and access to an unknown distribution of inputs, the goal is to minimize the expected loss with respect to the distribution. Assuming that the loss function is differentiable, the gradient of the loss serves as a gradient estimator (see \Cref{def:grad_estimator}) for the risk function. We refer the reader to \citet[Section 14.5.1]{SSS14} for a complete overview of SGD for risk minimization.

For the sake of completeness, we formulate Transcript Learning (TL, \Cref{alg:TL}) in the framework of Risk Minimization for Supervised Learning. This is not strictly needed for the proof of \Cref{thm:TL}, but is an illuminating connection. Although multiple loss functions may achieve our ultimate goal---learning Self-Proving models---in what follows we define the loss that corresponds to TL.

Fix a verifier $V$ and let $\Transcript^\ast_V$ denote a distribution over accepting transcripts. We define
\begin{equation}\label{eq:loss}
	\texttt{loss} \left( \theta, (x,\pi^\ast) \right) \defeq \Pr_{\pi \gets \Transcript^\theta_V(x)}\left[\pi \neq \pi^\ast\right],
\end{equation}
where $\pi^\ast$ and $\pi$ share the same verifier messages (as in \Cref{lem:grad_restated}) so the inequality is only over the prover's messages, namely $\Pr_{\pi \gets \Transcript^\theta_V(x)}\left[\pi \neq \pi^\ast\right] = 
\Pr_{\pi \gets \Transcript^\theta_V(x)}[(y,a_1,\dots,a_R) \neq (y^\ast,a_1^\ast,\dots,a_R^\ast)]$.\footnote{This loss is not to be confused with those discussed in \Cref{apx:defs-gen}. Here, we are simply explaining how TL can be viewed as a supervised risk minimizer for the loss function defined in \Cref{eq:loss}.}

The risk function is the expected value of the loss over the joint distribution of inputs and accepting transcripts $\mu \times \Transcript^\ast_V(\mu)$:
\[
\texttt{Risk} \left( \theta \right) \defeq \E_{\substack{x \gets \mu \\ \pi^\ast \gets \Transcript^\ast_V}} \left[ \texttt{loss} \left( \theta, (x,\pi^*) \right) \right],
\]
which means that the \emph{agreement function} defined in \Cref{thm:TL}:
	\begin{equation*}
		\agree(\theta) =  \Pr_{\substack{x \gets \mu \\ \pi^* \gets \Transcript^\ast_V(x) \\ \pi \gets \Transcript^\theta_V(x)}}\left[\pi = \pi^* \right],
	\end{equation*}
satisfies $\agree(\theta) = 1 - \texttt{Risk}(\theta)$.

Thus, maximizing the agreement is equivalent to minimizing the risk. The hypothesis class over which the optimization is performed is the ball of radius $\Bnorm$, i.e., $\left\{ \theta \in \R^d : ||\theta|| < \Bnorm \right\}$.  The assumption that $\agree$ is concave in $\theta$ implies that the loss function is convex in $\theta$, which is the required assumption for using SGD for risk minimization.

\subsubsection{\texorpdfstring{Proof of \Cref{thm:TL}}{Proof of Theorem 5.1}}\label{apx:actual-proof-of-TL}

Our strategy is to cast TL as Stochastic Gradient Ascent and apply \Cref{fact:sga}.
	Let $\eps$, $\Bnorm$, $\Blip$ and $\Cprov$ as in the theorem statement be given. Let $\theta^\ast$ be such that $A(\theta^\ast) \geq 1 - \eps/2$ and $||\theta^\ast|| \leq \Bnorm$.
	
	First, notice that 
	\begin{equation*}
		\E_{\bar{\theta}}\left[\ver_{V,\mu}(\bar{\theta})\right] \geq \E_{\bar{\theta}}[A(\bar{\theta})].
	\end{equation*}
	This holds because, for any $x$ and model $P_\theta$, whenever the transcript generated by $\Transcript^\theta(x)$ agrees with $\pi^\ast$, then the verifier accepts (because $\pi^\ast$ is honest). Therefore, to prove the theorem it suffices to show that
	\begin{equation*}
		\E_{\bar{\theta}}[A(\bar{\theta})] \geq 1 - \eps.
	\end{equation*}
	
	Following the notation in \Cref{alg:TL}, in every iteration $i \in [N]$ the norm of the update step is
	\begin{multline*}
		\left\lVert  \prod_{\substack{r \in [R]\cup \{0\} \\ s \in S(r)}} \alpha_{s}(\theta_i) \cdot
           		\sum_{\substack{r \in [R] \cup \{0\} \\ s \in S(r)}}\vec{d}_s(\theta_i) \right \rVert
           		= \left| \prod_{\substack{r \in [R] \cup \{0\}\\ s \in S(r)}} \alpha_{s}(\theta_i) \right| \cdot
           		\left\lVert \sum_{\substack{r \in [R] \cup \{0\} \\ s \in S(r)}}\vec{d}_s(\theta_i) \right \rVert
                    \\
           		\leq 1 \cdot  \sum_{\substack{r \in [R] \cup \{0\} \\ s \in S(r)}}\left\lVert\vec{d}_s(\theta_i) \right\rVert,
    	\end{multline*}
	where the inequality is because $\alpha_s(\theta_i)$ are probabilities, so $\leq1$. Moreover, we have
	\begin{equation*}
		\sum_{\substack{r \in [R] \cup \{0\} \\ s \in S(r)}}\left\lVert\vec{d}_s(\theta_i) \right\rVert\leq \sum_{\substack{r \in [R] \cup \{0\} \\ s \in S(r)}} \Blip \leq \Cprov \cdot \Blip.
	\end{equation*}
	The first inequality is by definition of $\Blip$ as an upper-bound on the gradient of $P_\theta$'s logits. The second is because, by definition, $\Cprov$ is an upper-bound on the number of tokens sent by the prover in the proof system, which is exactly the number of terms in the sum: $r$ indexes rounds, and $s$ indexes tokens sent in each round.

    To conclude, \Cref{lem:grad_restated} shows that TL samples from a gradient estimator for $A(\theta)$, while the above equation shows that the gradient is upper-bounded by $\Cprov \cdot \Blip$. We can therefore apply \Cref{fact:sga} to obtain
	\begin{equation*}
		\E_{\bar{\theta}}\left[\agree \left(\bar{\theta}\right)\right] \geq \agree(\theta^\ast) - \eps / 2 \geq (1 - \eps/2) - \eps/2 =  1 - \eps,
	\end{equation*}
	where the inequality is by definition of $\theta^\ast$. This completes the proof of \Cref{thm:TL}.

\section{Full Details of the Experimental Setup}
\subsection{The Bézout proof system for GCD} \label{app:verifying_the_GCD}
The Euclidean algorithm for computing the Greatest Common Divisor (GCD) of two integers is possibly the oldest algorithm still in use today \citep{Knuth69}. Its extended variant gives a simple proof system.

Before we dive in, let us clarify what we mean by \emph{a proof system for the GCD}. Prover Paul has two integers $212$ and $159$; he claims that $GCD(212,159)=53$. An inefficient way for Verifier Veronica to check Paul's answer is by executing the Euclidean algorithm on $(212,159)$ and confirm that the output is $53$. In an efficient proof system, Veronica asks Paul for a short string $\pi^\ast$ (describing two integers) with which she can easily compute the answer---without having to repeat Paul's work all over. On the other hand, if Paul were to claim that ``$GCD(212,159)=51$'' (it does not), then for any alleged proof $\pi$, Veronica would detect an error and reject Paul's claim.

The verifier in the proof system relies on the following fact.
\begin{fact}[Bézout's identity \citep{bezout}] \label{clm:bezout_identity}
    Let $x_0,x_1 \in \N$ and $z_0,z_1 \in \Z$. If $z_0 \cdot x_0 + z_1 \cdot x_1$ divides both $x_0$ and $x_1$,
    then $z_0 \cdot x_0 + z_1 \cdot x_1 = GCD(x_0,x_1)$.
\end{fact}
Any coefficients $z_0, z_1$ satisfying the assumption of \cref{clm:bezout_identity} are known as \emph{Bézout coefficients} for $(x_0,x_1)$. \Cref{clm:bezout_identity} immediately gives our simple proof system: For input $x=(x_0,x_1)$ and alleged GCD $y$, the honest prover sends (alleged) Bézout coefficients $(z_0,z_1)$. The Verifier accepts if and only if $y=z_0 \cdot x_0 + z_1 \cdot x_1$ and $y$ divides both $x_0$ and $x_1$.

In this proof system the Verifier does not need to make any query; to fit within \Cref{def:sys-ver}, we can have the verifier issue a dummy query. Furthermore, by \Cref{clm:bezout_identity} it is complete and has soundness error $s=0$. Lastly, we note that the Verifier only needs to perform two multiplications, an addition, and two modulus operations; in that sense, verification is more efficient than computing the GCD in the Euclidean algorithm as required by \cref{rem:verifier_efficiency}.

\paragraph{Annotations.} To describe how a proof $z=(z_0,z_1)$ is annotated, let us first note how it can be computed.
The Bézout coefficients can be found by an extension of the Euclidean algorithm. It is described in \Cref{alg:extended_euclidean}.\footnote{Our description follows \url{https://en.wikipedia.org/wiki/Extended_Euclidean_algorithm}.}

\begin{algorithm}[tb]
   \caption{Extended Euclidean algorithm}
    \label{alg:extended_euclidean}
    \KwInput{Nonzero integers $x_0, x_1 \in \N$.}
    \KwOutput{Integers $(y,z_0,z_1)$, such that $y = GCD(x_0,x_1)$ and $(z_0,z_1)$ are Bézout coefficients for $(x_0,x_1)$.}
    Initialize $r_0 = x_0$, $r_1 = x_1$, $s_0 = 1$, $s_1 = 0$, and $q = 0$. \\
     \While{$r_1 \neq 0$}{
         Update $q \defeq \lfloor r_{0} / r_{1} \rfloor$. \\
         Update $(r_0, r_1) \defeq (r_1, r_0 - q \times r_1)$.\\
         Update $(s_0, s_1) \defeq (s_1, s_0 - q \times s_1)$.
     }
     Output GCD $y = r_0$ and Bézout coefficients $z_0 \defeq s_0$ and $z_1 \defeq (r_0 - s_0 \cdot x_0) / x_1$.
\end{algorithm}

Referring to \Cref{alg:extended_euclidean}, the annotation of a proof $z=(z_0,z_1)$ will consist of intermediate steps in its computation. Suppose that in each iteration of the While-loop, the algorithm stores each of $r_0$, $s_0$ and $q$ in an arrays $\vec{r_0}$, $\vec{s_0}$ and $\vec{q}$. The annotation $\tilde{z}$ of $z$ is obtained by concatenating each of these arrays. In practice, to avoid the transformer block (context) size from growing too large, we fix a cutoff $T$ and first trim each array to its first $T$ elements.

We formalize this in the terminology of \Cref{apx:annot} by defining a Transcript Annotator and Answer Extractor. Note that, since our proof system consists only of one ``answer'' $z$ send from the prover to the verifier, the entire transcript $\pi$ is simply $z=(z_0,z_1)$. Since the verification is deterministic, this means that the proof system is of an $\mathsf{NP}$ type (however, note that the search problem of finding the ``$\mathsf{NP}$-witness'' $z=(z_0,z_1)$ is in fact in $\mathsf{P}$). 

\begin{itemize}
	\item \emph{Transcript Annotator $A$:} For a fixed cutoff $T$ and given input $x=(x_0,x_1)$ and transcript $z=(z_0,z_1)$, $A$ executes \Cref{alg:extended_euclidean} on input $x=(x_0,x_1)$. During the execution, $A$ stores the first $T$ intermediate values of $r_0$, $s_0$ and $q$ in arrays $\vec{r_0}$, $\vec{s_0}$ and $\vec{q}$. It outputs $A(x,z) \defeq (\vec{r_0}, \vec{s_0}, \vec{q}, z)$.
	\item \emph{Answer Extractor $E$:} Given an annotated transcript $\tilde{z} = (\vec{r_0}, \vec{s_0}, \vec{q}, z)$, outputs $E(\tilde{z}) \defeq z$.
\end{itemize}
We note that the computational complexity of $A$ is roughly that of the honest prover, i.e., \Cref{alg:extended_euclidean} (up to additional space due to storing intermediate values). As for $E$, it can be implemented in logarithmic space and linear running time in $|\tilde{z}|$, i.e., the length of the description.\footnote{That is, if integers are represented by $n$-bits, then $E$ has space complexity $O(\log n + \log T)$ and running time $O(n\cdot T)$.}

\subsection{Implementation details} \label{apx:experiments-details}
Code, data and models are available at \url{\codeurl}.

\paragraph{Model architecture.} We use Karpathy's \emph{nanoGPT}\footnote{\url{https://github.com/karpathy/nanoGPT}.} implementation of GPT. Note that we train the model ``from scratch'' only on sequences related to the GCD problem, rather than starting from a pretrained checkpoint.
We use a 6.3M parameter architecture of $8$ layers, $8$ attention heads, and $256$ embedding dimensions. We optimized hyperparameters via a random hyperparameter search, arriving at learning rate $0.0007$, AdamW $\beta_1=0.733$ and $\beta_2=0.95$, $10\%$ learning rate decay factor, no dropout, gradient clipping at $2.0$, no warmup iterations, and $10\%$ weight decay.

\paragraph{Data.} We sample integers from the $\log_{10}$-uniform distribution over $\{1,\dots,10^4\}$. Models in \Cref{tab:main,fig:annot} are trained for 100K iterations on a dataset of $\approx$10M samples. For \Cref{fig:bases} (base ablation) we train for 20K iterations on a dataset of $\approx$1M samples; this is because this setting required 68 many runs in total, whereas the annotation-cutoff ablation required 18 longer runs.

\paragraph{Compute.} All experiments were run on a machine with an NVIDIA A10G GPU, 64GB of RAM, and 32 CPU cores. The longest experiment was the single RLVF run, which took one month and four days. The annotation-cutoff ablation runs took about 75 minutes each. Base of representation ablation runs were shorter at about 15 minutes each. The total running time of the Transcript Learning experiments was approximately 40 hours (excluding time dedicated to a random hyperparameter search), and the RLVF experiment took another month and four days. The overall disk space needed for our models and data is 4GB.

\paragraph{Representing integers.} We fully describe how integer sequences are encoded. As a running example, we will use base $210$. To encode a sequence of integers, each integer is encoded in base $210$, a sign is prepended and a delimiter is appended, with a unique delimiter identifying each component of the sequence. For example, consider the input integers $x_0=212$ (which is $12$ in base 210) and $x_1=159$. Their GCD is $y=53$, with Bézout coefficients $z_0=1$ and $z_1=-1$. Therefore, the sequence $(212,159,53,1,-1)$ is encoded as
\begin{center}
    \texttt{+,1,2,x0,+,159,x1,+,53,y,+,1,z0,-,1,z1}
\end{center}
where commas are added to distinguish between different tokens. Null tokens are appended to pad all sequences in a dataset to the same length. Both the input and the padding components are ignored when computing the loss and updating parameters.

\paragraph{Annotations} Annotations are encoded as above, with each component in an intermediate step $\pi_t$ delimited by a unique token. Since different integer pairs may require a different number of intermediate steps to compute the Bézout coefficients, we chose to pad all annotations to the same length $T$ by the last step $\pi_T$ in the sequence (which consists of the final Bézout coefficients). This ensures that the final component output by the model in each sequence should be the Bézout coefficient, and allows us to batch model testing (generation and evaluation) resulting in a 1000x speed-up over sequential testing.

As an example, consider the inputs $x_0 = 46$ and $x_1 = 39$. Tracing through the execution of \Cref{alg:extended_euclidean}, we have
\[
\begin{array}{c|c|c|c|c|c|c|c|c}
 & x_0 & x_1 & y & \vec{s_0} & \vec{r_0} & \vec{q} & z_0 & z_1 \\
\hline
 & 46 & 39 &  & 1 & 46 & 1 &  & \\
 &  &  &  & 0 & 39 & 5 &  & \\
 &  &  &  & 1 & 7 & 1 &  & \\
 &  &  &  & -5 & 4 & 1 &  & \\
 &  &  &  & 6 & 3 & 3  &  &\\
 &  &  & 1 &  &  & & -11 & 13\\
\end{array}
\]

To encode this as an annotated transcript for the transformer, we must specify a base of representation and an annotation cutoff. Suppose that we wish to encode this instance in base $B=10$ and cutoff $T=3$. Then the input with the annotated transcript is encoded as
\begin{center}
\texttt{+,4,6,x0,+,3,9,x1,+,1,y,}\\
\texttt{+,1,z0',+,4,6,z1',+,1,q',}\\
\texttt{+,0,z0'',+,3,9,z1'',+,5,q'',}\\
\texttt{+,1,z0''',+,7,z1''',+,1,q''',}\\
\texttt{-,1,1,z0,+,1,3,z1}
\end{center}
where commas are used to separate between tokens, and linebreaks are added only for clarity. Notice the three types of tokens: signs, digits, and delimiters. Notice also that the output $y$ is added immediately after the input, followed by the annotated transcript (whose six tokens comprise the proof itself). Since the Self-Proving model we train has causal attention masking, placing the output $y$ before the proof means that the model ``commits'' to an output and only then proves it.
\end{document}